\documentclass[pdflatex,sn-mathphys-num]{sn-jnl}

\usepackage{graphicx}%
\usepackage{multirow}%
\usepackage{amsmath,amssymb,amsfonts}%
\usepackage{amsthm}%
\usepackage{mathrsfs}%
\usepackage[title]{appendix}%
\usepackage{xcolor}%
\usepackage{textcomp}%
\usepackage{manyfoot}%
\usepackage{booktabs}%
\usepackage{algorithm}%
\usepackage{algorithmicx}%
\usepackage{algpseudocode}%
\usepackage{listings}%

\usepackage{booktabs}
\usepackage{subcaption}
\usepackage{enumitem}
\usepackage{makecell}
\usepackage{placeins}

\AtBeginDocument{%
    \let\boldmath\unboldmath
}

\theoremstyle{thmstyleone}%

\theoremstyle{thmstyletwo}%

\theoremstyle{thmstylethree}%

\begin{document}
\title[Toward Clinical Translation of Intracranial 4D Flow MRI Super-Resolution]{Toward Clinical Translation of Deep Learning-Based Intracranial Super-Resolution 4D Flow MRI Using Multi-Cohort Transfer Learning}

\author*[1,2]{\fnm{Oliver} \sur{Welin Odeback}}\email{oliver.welin.odeback@ki.se}
\author[2]{\fnm{Susanne} \sur{Schnell}}
\author[2]{\fnm{Patrick} \sur{Winter}}
\author[3,4]{\fnm{Johan} \sur{Lundberg}}
\author[3]{\fnm{Fredrik} \sur{Piehl}}
\author[1,3,4]{\fnm{Tobias} \sur{Granberg}}
\author[1,5]{\fnm{Alexander} \sur{Fyrdahl}}
\author[1,6]{\fnm{David} \sur{Marlevi}}

\affil*[1]{\orgdiv{Department of Molecular Medicine and Surgery}, \orgname{Karolinska Institutet}, \orgaddress{\city{Stockholm}, \country{Sweden}}}
\affil[2]{\orgdiv{Department of Medical Physics}, \orgname{University of Greifswald}, \orgaddress{\city{Greifswald}, \country{Germany}}}
\affil[3]{\orgdiv{Department of Clinical Neuroscience}, \orgname{Karolinska Institutet}, \orgaddress{\city{Stockholm}, \country{Sweden}}}
\affil[4]{\orgdiv{Department of Neuroradiology}, \orgname{Karolinska University Hospital}, \orgaddress{\city{Stockholm}, \country{Sweden}}}
\affil[5]{\orgdiv{Department of Clinical Physiology}, \orgname{Karolinska University Hospital}, \orgaddress{\city{Stockholm}, \country{Sweden}}}
\affil[6]{\orgdiv{Institute for Medical Engineering and Science}, \orgname{Massachusetts Institute of Technology}, \orgaddress{\city{Cambridge}, \state{MA}, \country{USA}}}

\abstract{

\textbf{Purpose:}
To evaluate the \textit{in vivo} potential and cross-cohort generalization of deep learning-based super-resolution for intracranial 4D Flow MRI, assessing to what extent transfer learning improves regional accuracy across heterogeneous neurological cohorts.

\textbf{Materials and Methods:}
This mixed prospective-retrospective study included 12 prospectively enrolled participants (March 16–November 19, 2025) and 6 retrospective healthy extracranial participants (August 26–September 24, 2024). Eighteen participants (mean age, 35.4 years; range, 23-70 years; 8 women) were analyzed: 12 healthy participants, 3 with moyamoya disease, and 3 with multiple sclerosis. A convolutional neural network pretrained on synthetic cerebrovascular data was fine-tuned using cohort-specific or multi-cohort data. Paired 1.2-mm and 0.6-mm 4D Flow MRI scans were acquired. High-resolution-derived 1.2-mm inputs were used for fine-tuning and voxel-wise evaluation; native standard-resolution scans were used for angiographic and hemodynamic analyses. Outcomes included velocity-normalized root mean square error (vNRMSE), Bland–Altman agreement, precision–recall area under the curve (PR-AUC), and hemodynamic measures. 

\textbf{Results:}
Synthetic pretraining reduced mean artery-wise vNRMSE versus interpolation (0.147 vs 0.166). Pooled transfer learning yielded the lowest mean vNRMSE (0.119 vs 0.127 for cohort-specific models and 0.147 for synthetic pretraining) and narrower 95\% limits-of-agreement widths than synthetic pretraining across all test cases (0.109--0.222 vs 0.154--0.278 m/s). Super-resolved phase-contrast angiography had higher PR-AUC than native standard-resolution imaging in all cohorts (0.848--0.896 vs 0.486--0.805).

\textbf{Conclusion:}
Pooled multi-cohort transfer learning reduced reconstruction error and narrowed voxel-wise limits of agreement relative to synthetic pretraining. Super-resolution also improved vessel depiction relative to both standard- and high-resolution imaging.
}


\maketitle

\section{Introduction}\label{sec1}

Traditionally, neurovascular disease has been diagnosed from an anatomical paradigm, with imaging techniques primarily used to qualitatively \textit{visualize} intracranial vascular morphology and pathology. Yet many clinical questions are fundamentally hemodynamic, including whether flow is compromised distal to a stenosis, whether collateral supply is sufficient, or how flow redistributes after revascularization. Quantitative clinical intracranial hemodynamic assessment, however, remains challenging, with the shielding cranium, limited interrogation windows, and tortuous vascular anatomy restricting the use of routine tools used in other vascular compartments~\cite{naqvi2013transcranial, morgan20214d}. In this context, four-dimensional flow magnetic resonance imaging (4D Flow MRI) enables non-invasive visualization and quantification of cerebrovascular blood flow across a large vascular territory~\cite{schnell2016four, waahlin20224d}. However, despite translational advances in acquisition, correction, and quantification~\cite{schnell2017accelerated, ma2020efficient, aristova2022accelerated, rivera2022motion, wen2019test}, a fundamental trade-off remains between spatial resolution, signal-to-noise ratio, and scan time: high-resolution acquisitions are lengthy and impractical, whereas standard-resolution acquisitions cannot capture small branches, underestimate peak velocities, and reduce vessel conspicuity. These challenges are particularly relevant in steno-occlusive disease such as moyamoya, where hemodynamic assessment depends on resolving altered flow in small collateral vessels and the downstream circulation~\cite{matsushige2018visualization, yu2023analyzing}.

To address this, deep learning-based super-resolution has emerged as a practical strategy to increase image resolution without extending scan time or modifying hardware or pulse sequences, with the potential to sharpen vessel depiction and improve hemodynamic quantification post-scan~\cite{ferdian2023cerebrovascular, shit2022srflow}. However, existing intracranial super-resolution models are predominantly trained on synthetic surrogate data, with \textit{in vivo} validation generally limited to small cohorts or indirect reference settings~\cite{ferdian2023cerebrovascular, shit2022srflow, ericsson2024generalized}. Specifically, it remains unclear whether synthetic training generalizes sufficiently to \textit{in vivo} scans - where noise characteristics, anatomy, and disease-related flow patterns differ - or whether additional adaptation is required. In this context, \textit{transfer learning} offers a potential mechanism for adapting synthetic-pretrained models to \textit{in vivo} data, supported by extensive evidence from other imaging domains~\cite{valverde2021transfer}. However, the benefit of such fine-tuning for intracranial super-resolution 4D Flow MRI, particularly across heterogeneous clinical cohorts, has not been established.

The aim of this study was therefore to evaluate the \textit{in vivo} performance and cross-cohort generalization of a synthetic-pretrained super-resolution model for intracranial 4D Flow MRI using paired standard- and high-resolution acquisitions, and to determine whether transfer learning provides additional benefit across clinically relevant neurological cohorts. We hypothesized that transfer learning would improve performance on acquired \textit{in vivo} data, reflected by reduced reconstruction error relative to synthetic pretraining alone. We further evaluated angiographic vessel depiction and quantitative hemodynamic performance to assess the clinical potential of super-resolution across diverse cerebrovascular settings.

\section{Materials and methods}\label{sec2}

\begin{figure}[h]
\centering
\makebox[\textwidth][c]{%
    \includegraphics[width=1.4\textwidth]{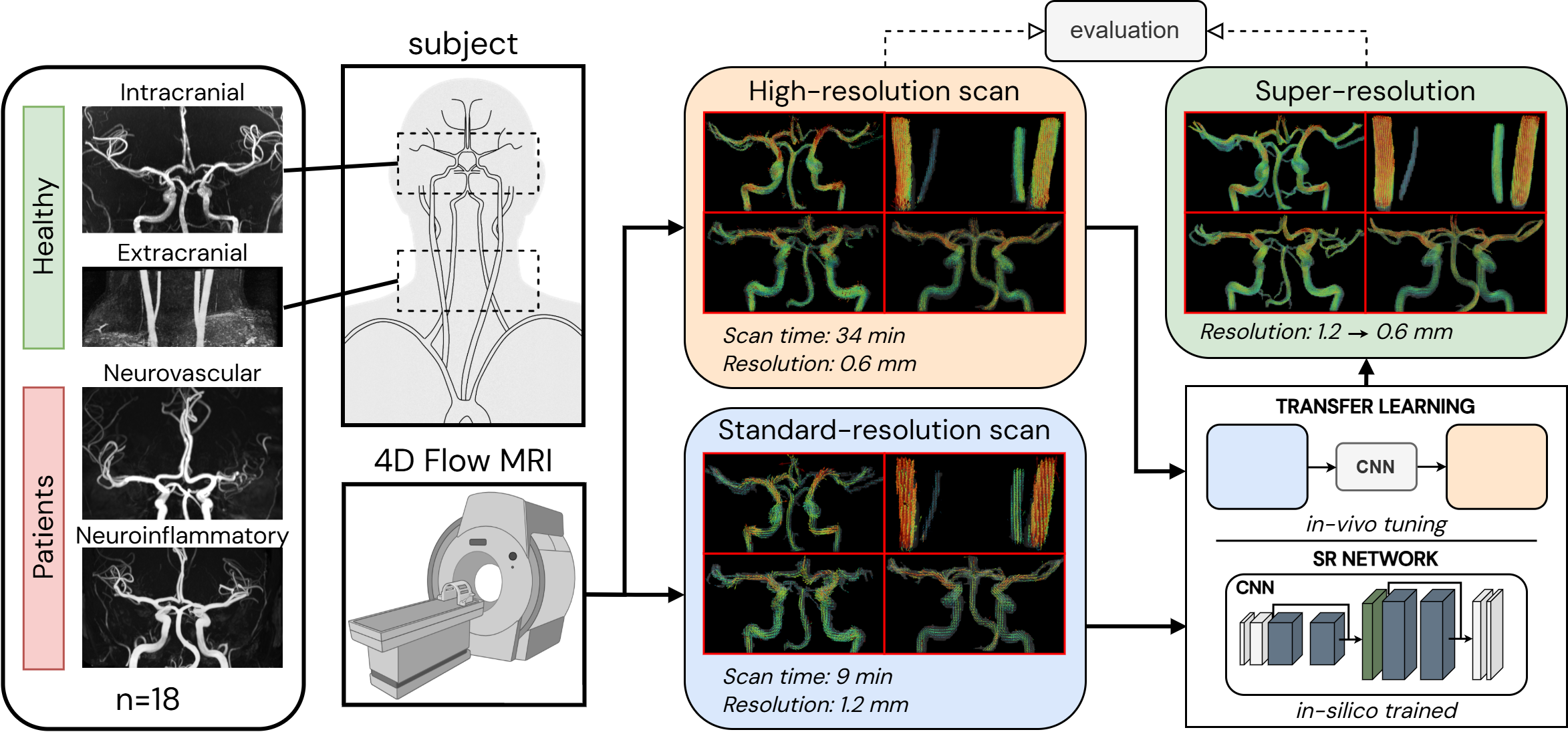}
}
\caption{Methodological overview. Paired standard- and high-resolution four-dimensional flow magnetic resonance imaging (4D Flow MRI) scans were acquired in healthy volunteers and patients ($n=18$). A synthetic-pretrained convolutional neural network (CNN) was fine-tuned using \textit{in vivo} data via transfer learning to super-resolve standard-resolution data, which were evaluated against the corresponding high-resolution reference scans.
}
\label{fig1}
\end{figure}

\subsection{Study Design and Participants}

This mixed prospective-retrospective study evaluated the \textit{in vivo} generalization of a synthetic-pretrained super-resolution model and whether transfer learning improved performance across heterogeneous cohorts. The study included 18 participants (8 women, 10 men; mean age, 35.4 years; range, 23--70 years) in four cohorts: healthy intracranial ($n=6$), healthy extracranial ($n=6$), neurovascular (moyamoya, $n=3$), and neuroinflammatory (multiple sclerosis, $n=3$). An overview of the study and evaluation workflow is provided in Figure~\ref{fig1}. The intracranial cohorts were prospectively enrolled between March 16, 2025 and November 19, 2025. The extracranial cohort was retrospectively selected from examinations acquired between August 26, 2024 and September 24, 2024. Healthy participants were recruited as volunteers. Eligibility for the patient cohorts required a clinical diagnosis of moyamoya disease or multiple sclerosis, respectively, and no contraindications for MRI. Participants were enrolled based on availability during the study period, and all datasets meeting the study criteria were included. No formal power calculation was performed; the sample size was based on the available paired acquisitions and the feasibility of participant-level training and held-out test partitions.

The study was approved by the Swedish Ethical Review Authority (EPM, 2023-07667-01). Written informed consent was obtained from all enrolled participants. Data were pseudonymized before analysis, with direct personal identifiers removed and participants represented by study-specific identifiers. No participant overlapped with prior publications. Participant characteristics, including sex-stratified age summaries, are provided in Table~\ref{tab:demographics}.

\begin{table}[t]
\centering
\footnotesize
\setlength{\tabcolsep}{3.5pt}
\caption{Characteristics of study cohorts.}
\label{tab:demographics}

\begin{tabular}{lccccc}
\toprule
& Overall
& Intracranial
& Extracranial
& \shortstack{Neuro-\\vascular}
& \shortstack{Neuro-\\inflammatory} \\
\midrule

Participants, n
& 18 & 6 & 6 & 3 & 3 \\

Age, y$^{*}$
& 35.4 (23--70)
& 35.8 (25--70)
& 30.2 (23--38)
& 36.3 (31--41)
& 44.0 (34--59) \\

Women, n
& 8 & 2 & 2 & 3 & 1 \\

Men, n
& 10 & 4 & 4 & 0 & 2 \\

BMI$^{\dagger\,\ddagger}$
& 22.6 (19.8--32.3)
& 24.4 (19.8--28.7)
& 22.5 (20.6--32.3)
& 22.5 (21.3--24.1)
& 21.4 (21.2--24.8) \\

Heart rate, beats/min$^{\ddagger}$
& 60.8 (49.3--71.5)
& 60.8 (50.0--68.3)
& 58.2 (49.3--61.8)
& 67.0 (60.0--71.5)
& 61.5 (60.0--65.7) \\

Site
& -- & Site 1 & Site 2 & Site 1 & Site 1 \\

Disease
& -- & -- & -- & Moyamoya & Multiple sclerosis \\

\bottomrule
\end{tabular}

\begin{flushleft}
\footnotesize
$^{*}$Age is reported as mean (range). Overall age by sex:
women, 38.3 (23--70) years; men, 33.1 (25--59) years. \\
$^{\dagger}$BMI is calculated as weight (kg) divided by height (m)$^{2}$. \\
$^{\ddagger}$BMI and heart rate are reported as median (range).
\end{flushleft}
\end{table}

\subsection{MRI Acquisition and Image Processing}

All participants underwent dual-velocity-encoding, kt-GRAPPA-accelerated 4D Flow MRI at 3~T (acceleration factor $R=5$), with paired standard-resolution (LR; 1.2-mm isotropic) and high-resolution (HR; 0.6-mm isotropic) acquisitions and 3D time-of-flight (TOF) MR angiography. For the intracranial cohorts, the imaging field of view was centered on the Circle of Willis, whereas the extracranial cohort was acquired with the field of view centered on the cervical carotid arteries. Intracranial and extracranial imaging were performed on two scanner platforms at separate sites. Acquisition parameters are summarized in Table~\ref{tab:acq_params}.

\begin{table*}[b]
\centering
\footnotesize
\caption{MRI acquisition parameters.}
\label{tab:acq_params}

\begin{tabular}{lcccc}
\toprule
& \multicolumn{2}{c}{Site 1}
& \multicolumn{2}{c}{Site 2} \\
\cmidrule(lr){2-3}
\cmidrule(lr){4-5}

Parameter & LR & HR & LR & HR \\
\midrule

Resolution (mm)
& 1.2 & 0.6 & 1.2 & 0.6 \\

TE (ms)
& 4.5 & 4.96 & 3.08 & 3.72 \\

TR (ms)
& 100.8 & 103.6 & 79.8 & 92.4 \\

Cardiac phases
& 5--8 & 5--8 & 11--12 & 9--11 \\

Flip angle ($^\circ$)
& 15 & 15 & 15 & 15 \\

Scanner (3 T)
& Prisma Fit & Prisma Fit & Magnetom Vida & Magnetom Vida \\

Low VENC (cm/s)
& 60 & 60 & 60 & 60 \\

High VENC (cm/s)
& 120 & 120 & 120 & 120 \\

Scan time (mm:ss)$^{*}$
& 9:32 [8:00--12:53]
& 37:20 [29:32--49:00]
& 8:42 [7:24--10:32]
& 31:51 [27:14--35:49] \\

\bottomrule
\end{tabular}
\vspace{-1.6em}
\begin{flushleft}
\footnotesize
LR = standard-resolution data; HR = high-resolution data;
TE = echo time; TR = repetition time;
VENC = velocity encoding. \\
$^{*}$Scan time is reported as median [range].
\end{flushleft}

\end{table*}

4D Flow MRI was processed using a standardized semiautomatic workflow in MATLAB (MathWorks), as described previously~\cite{vali2019semi, aristova2019standardized, winter2024automated}. A single TOF-derived vessel segmentation was generated per participant and rigidly registered (SPM12~\cite{spm12}) separately to the native LR and HR acquisitions. Data preprocessing, vessel labeling, centerline extraction, analysis-plane placement, and direction-consistency mask refinement are detailed in the Supplemental Material, Section~S1.

\subsection{Super-Resolution Model and Transfer Learning}

Synthetic pretraining used four patient-specific cerebrovascular computational fluid dynamics (CFD) velocity fields converted to paired synthetic low- and high-resolution 4D Flow MRI using an acquisition-mimicking pipeline incorporating k-space truncation, noise injection, and dual-velocity-encoding reconstruction. A residual 3D convolutional neural network (CNN) received three velocity components and magnitude information and predicted super-resolved velocity fields. Both synthetic pretraining and \textit{in vivo} fine-tuning used a region-weighted mean-square-error training objective that included a non-fluid objective to suppress extravascular predictions. Synthetic data generation, network architecture, and training details are provided in Supplemental~S2.

Five transfer-learned model variants were trained: four cohort-specific models and one pooled multi-cohort model. HR-derived LR inputs generated by k-space cropping were used for supervised fine-tuning and voxel-wise testing to ensure exact spatial and temporal correspondence. Native standard-resolution acquisitions were used for angiographic and hemodynamic analyses. For each cohort, one participant was held out for final testing. In the two healthy cohorts ($n=6$ each), four participants were used for training and one additional participant was used for participant-independent validation. In the two patient cohorts ($n=3$ each), one participant was held out for testing and the remaining two were used for training. Since no separate validation participant was available, validation was performed at the patch level by re-sampling patches from one of the training participants. Validation data were used only to monitor fine-tuning performance and select model weights, whereas all final evaluations were performed on the held-out test participants. The pooled multi-cohort model used the combined training and validation data from all cohorts and was evaluated on the same held-out test participants as the cohort-specific models. Detailed partitions, patch sampling, and model-selection procedures are provided in Supplemental~S3. Code for model training and evaluation, together with trained model weights, will be made publicly available upon publication.

\subsection{Outcome Measures and Evaluation}

The primary model-performance outcome was artery-wise voxel-wise reconstruction error between super-resolved velocities and the matched HR reference, quantified using velocity-normalized root mean square error (vNRMSE). Secondary outcomes included Bland--Altman agreement at peak systole, phase-contrast MR angiography (PCMRA) vessel depiction relative to the co-registered TOF-derived anatomical reference using precision--recall area under the curve (PR-AUC) and maximum Dice coefficient ($Dice_{\max}$), and comparisons against native-acquisition hemodynamic measures, including flow rate ($Q$), 95th-percentile velocity ($v_{p95}$), voxel-wise velocity distributions, and mean absolute divergence. The publicly available cerebrovascular 4DFlowNet model was applied to the same held-out data as an external model benchmark. 

Metric definitions and benchmark details are provided in Supplemental~S4--S5, with complementary linear regression described in Supplemental~S6. Because one participant per cohort was held out for testing and voxel- and plane-level observations were nested within participants, analyses were descriptive. No inferential hypothesis tests were performed and no $P$-value threshold was defined.

\section{Results}

All 18 enrolled participants were included in the analysis, with the four predefined cohorts each contributing one participant-disjoint test case. The remaining participants were assigned to training and validation partitions as described in Supplemental~S3.

\subsection{Transfer learning improves agreement with high-resolution reference}
\label{sec_VWP}

\begin{figure}[t!]
\centering
\includegraphics[width=1.0\textwidth]{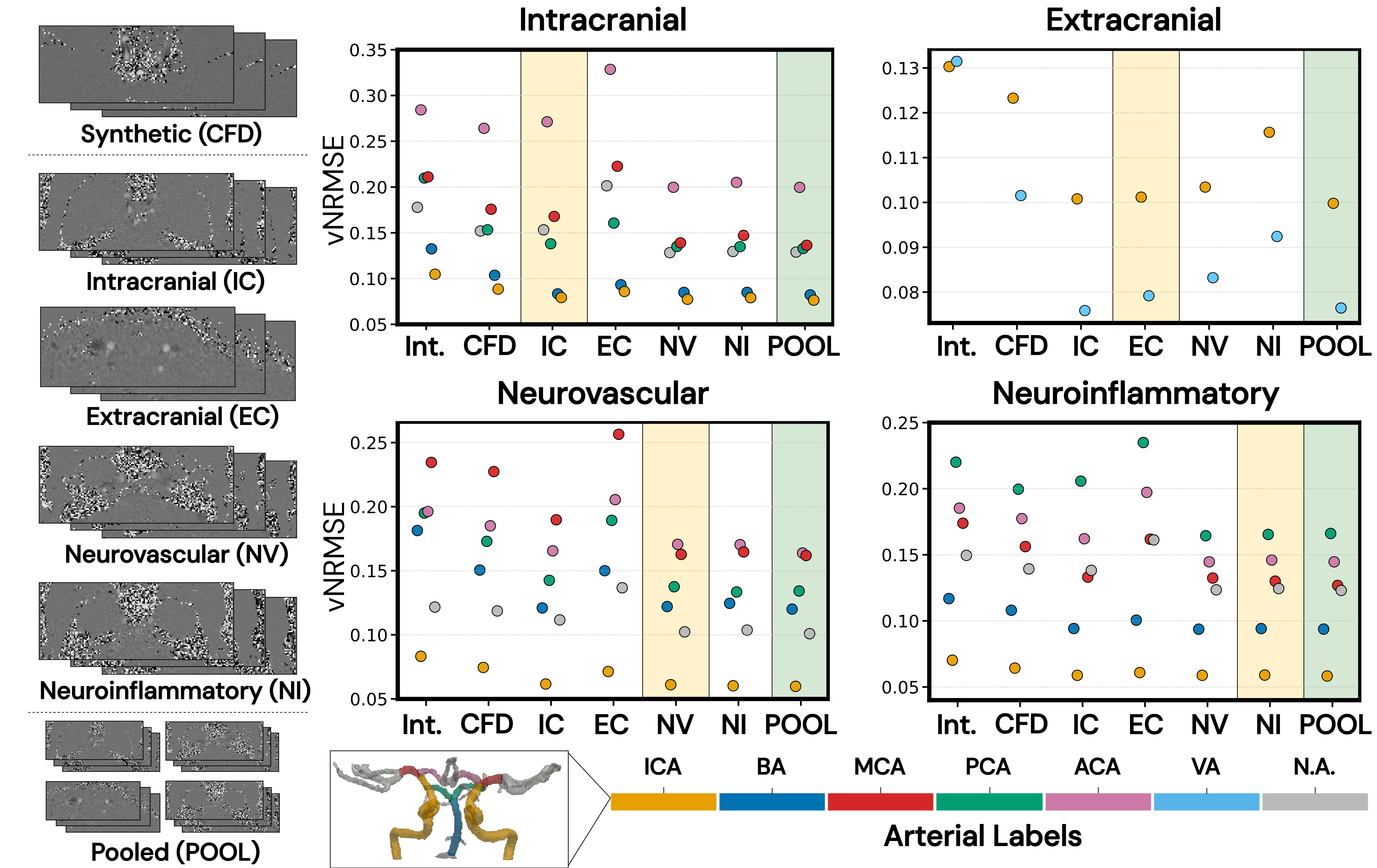}
\caption{Artery-wise voxel-wise reconstruction error relative to high-resolution (HR) 4D Flow MRI, quantified using velocity-normalized root mean square error (vNRMSE) and averaged over the full cardiac cycle. Results are shown for bicubic interpolation (Int.), the CFD-pretrained baseline (CFD), cohort-specific fine-tuned models (IC, EC, NV, NI), and the pooled multi-cohort model (POOL) across all 4D Flow MRI test sets. Yellow shaded regions indicate the cohort-specific fine-tuned model corresponding to the test domain in each panel. Green shaded regions indicate POOL. Point colors denote labeled arteries.
}
\label{fig2}
\end{figure}

\begin{figure}[t!]
\centering
\includegraphics[width=1.0\textwidth]{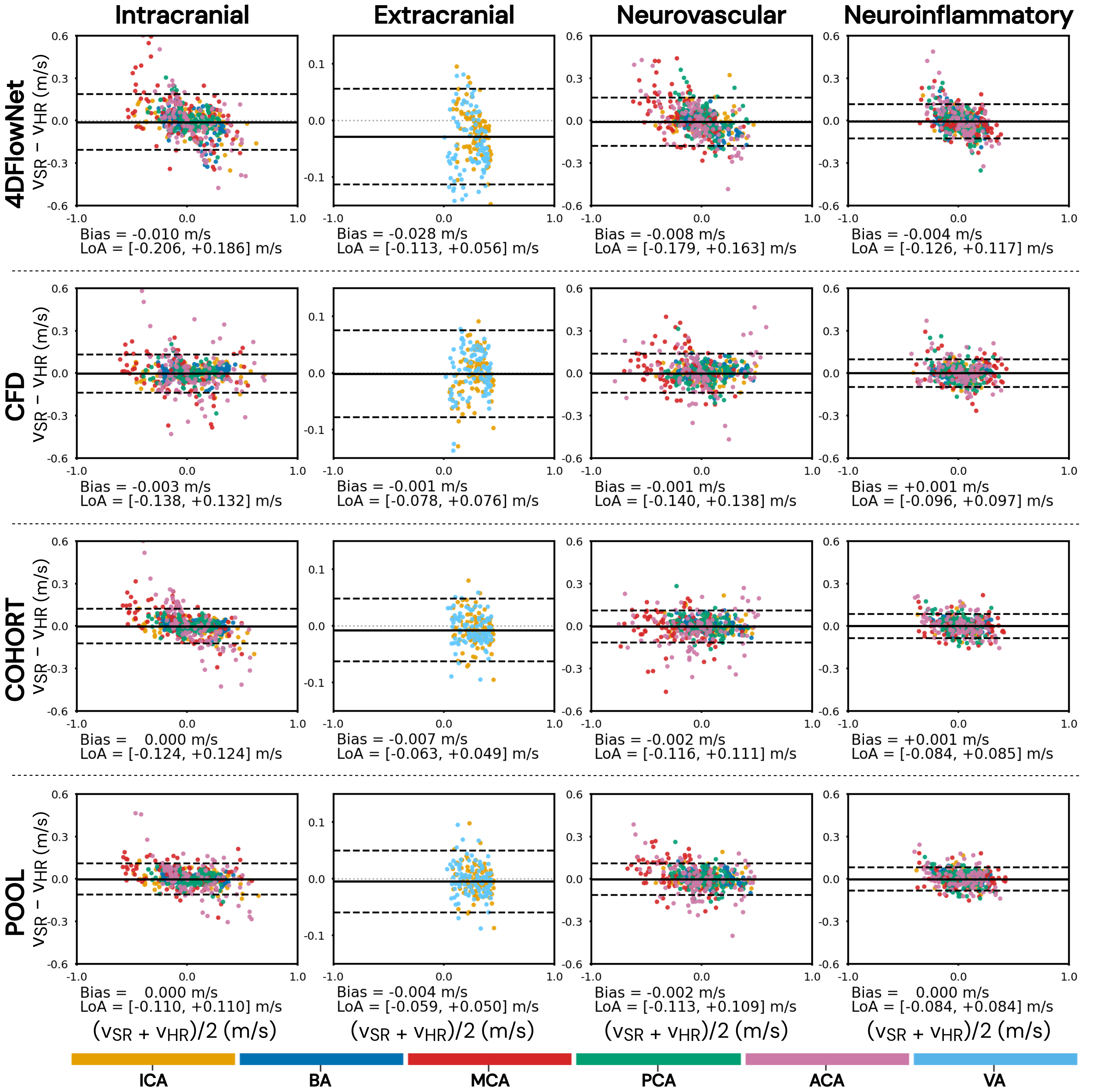}
\caption{Voxel-wise Bland--Altman analysis between high-resolution (HR) reference velocities (\(V_{\mathrm{HR}}\)) and super-resolved velocities (\(V_{\mathrm{SR}}\)) at peak systole. Rows show 4DFlowNet~\cite{ferdian2023cerebrovascular}, the CFD-pretrained baseline (CFD), the cohort-specific fine-tuned model (COHORT), and the pooled fine-tuned model (POOL); columns show the four test cases. For the intracranial, neurovascular, and neuroinflammatory cases, signed velocity components (\(V_x,V_y,V_z\)) were pooled. For the extracranial case, only \(V_z\) was analyzed, representing the predominant velocity component. Differences were defined as \(V_{\mathrm{SR}}-V_{\mathrm{HR}}\) and plotted against the corresponding pairwise mean. Solid horizontal lines indicate mean bias and dashed lines the 95\% limits of agreement (mean bias \(\pm1.96\) standard deviation), calculated using all analyzed voxels within the vessel mask. For visualization, an equal number of voxels per artery was sampled. Points are colored according to artery label.
}
\label{fig3}
\end{figure}

In Figure~\ref{fig2}, voxel-wise prediction performance is summarized across labeled arteries and averaged over the cardiac cycle for each cohort and model. Across cohorts, the CFD-pretrained model consistently improved on interpolation (mean aggregate vNRMSE: 0.147 vs 0.166), indicating that the synthetic-pretrained baseline retained transferable performance on acquired \textit{in vivo} data. Transfer learning further reduced vNRMSE in most arteries, with the pooled multi-cohort model yielding the lowest mean vNRMSE (0.119 vs 0.127 for the matched cohort-specific models and 0.147 for CFD). Cohort-specific transfer learning generally showed strong within-cohort performance but less consistent cross-cohort transfer. For example, the extracranial-tuned model reduced mean vNRMSE relative to CFD in the extracranial test case (0.090 vs 0.112) but performed less favorably when transferred to intracranial test cohorts (0.168 vs 0.151). In contrast, the pooled model showed consistent performance across the four test cases (mean vNRMSE: 0.119). Performance also varied by anatomical location, with larger arteries generally exhibiting lower errors (e.g., ICA and BA; mean vNRMSE: 0.082) as compared to smaller distal branches (e.g., MCA, ACA and PCA; mean vNRMSE: 0.152).

Figure~\ref{fig3} summarizes voxel-wise agreement between the HR reference and predicted super-resolved velocities at peak systole. Mean bias was close to zero for the CFD-pretrained and transfer-learned models across all four test cases (CFD: $-0.003$ to $0.001$ m/s; cohort-specific models (COHORT): $-0.007$ to $0.001$ m/s; pooled model (POOL): $-0.004$ to $0.000$ m/s). The 95\% limits of agreement narrowed after transfer learning, with POOL yielding narrower limits than CFD in all four test cases (width, $0.109$--$0.222$ vs $0.154$--$0.278$ m/s) and slightly narrower limits than COHORT ($0.112$--$0.248$ m/s). For comparison, 4DFlowNet showed larger negative biases ($-0.028$ to $-0.004$ m/s) and wider limits of agreement (width, $0.169$--$0.392$ m/s). Complementary linear regression analysis (Supplemental~S6) showed strong voxel-wise linear association for both COHORT ($R^2=0.906$--$0.929$) and POOL ($R^2=0.908$--$0.942$), although fitted slopes remained below unity (COHORT: $k=0.879$--$0.930$; POOL: $k=0.886$--$0.916$), indicating residual compression of the predicted velocity range.

\subsection{Super-resolution improves angiographic vessel delineation}

\begin{figure}[t!]
\centering
\includegraphics[width=1.0\textwidth]{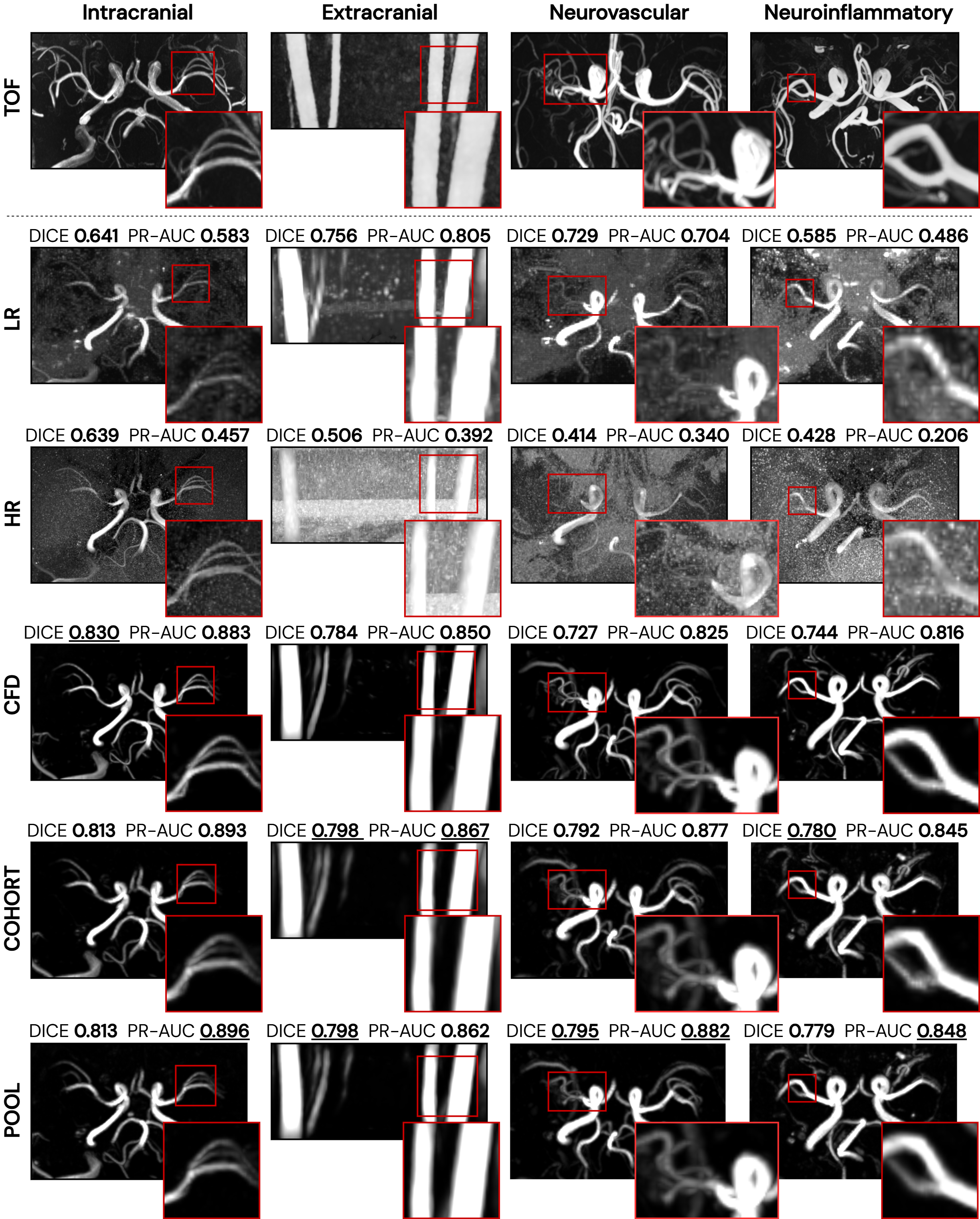}
\caption{Derived phase-contrast MR angiography (PCMRA) for the test subjects from each cohort. Rows show time-of-flight (TOF) angiography as anatomical reference, with PCMRA derived from acquired standard-resolution (LR) and high-resolution (HR) 4D Flow MRI, and PCMRA from the CFD-pretrained baseline (CFD), cohort-specific fine-tuned (COHORT), and pooled fine-tuned (POOL) super-resolution models. For each PCMRA image, the area under the precision--recall curve (PR-AUC) and maximum Dice coefficient (DICE) with respect to the corresponding TOF-derived vessel mask are reported. The highest PR-AUC and $Dice_{\max}$ values for each cohort are underlined.
}
\label{fig4}
\end{figure}

In Figure~\ref{fig4}, PCMRA visualizations derived from LR, HR, and super-resolved reconstructions across all four cohorts are compared with the corresponding TOF angiography as anatomical reference. Consistent with the background-suppression objective incorporated during training, all three super-resolution models showed reduced background signal relative to LR and HR, with visually clearer delineation of major arterial branches. In the neuroinflammatory case, the M2 branches were faint on LR and HR PCMRA but more clearly visualized on the super-resolved images. In the neurovascular case, the carotid siphon and downstream branches were likewise more apparent after super-resolution. Smaller vessels surrounding the affected proximal M1 were also more apparent on super-resolved PCMRA than on LR and HR, although their depiction remained incomplete relative to TOF.

The qualitative findings were supported by quantitative comparison with the TOF-derived vessel masks. PR-AUC was higher for every super-resolved model than for native LR or HR in all four test cases (LR: $0.486$--$0.805$; HR: $0.206$--$0.457$; CFD: $0.816$--$0.883$; COHORT: $0.845$--$0.893$; POOL: $0.848$--$0.896$). Corresponding $Dice_{\max}$ values were $0.585$--$0.756$ for LR, $0.414$--$0.639$ for HR, $0.727$--$0.830$ for CFD, $0.780$--$0.813$ for COHORT, and $0.779$--$0.813$ for POOL.

\subsection{Super-resolution preserved quantitative hemodynamics}

\begin{figure}[p]
\centering
\includegraphics[width=\textwidth,height=0.85\textheight,keepaspectratio]{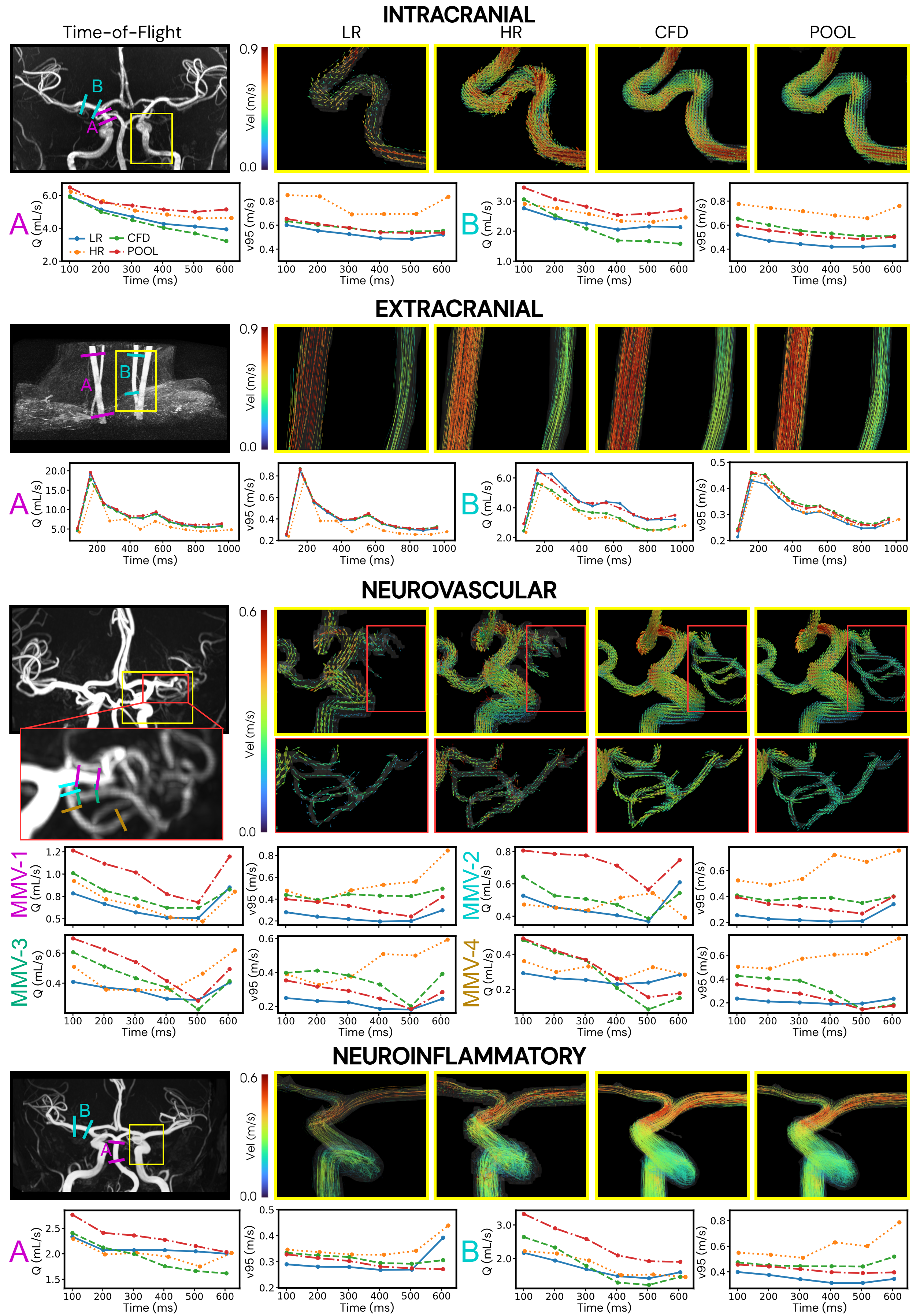}
\caption{
Plane-based hemodynamic comparison across all test subjects. For each cohort, the time-of-flight angiogram indicates the analyzed vessel regions, with colored markers denoting the first and last analysis plane of selected plane groups. Velocity vector fields (Intracranial, Neurovascular) or streamlines (Extracranial, Neuroinflammatory) at peak systole are shown for standard-resolution (LR), high-resolution (HR), CFD-pretrained (CFD), and pooled fine-tuned (POOL) within the highlighted region. For visual depiction of SR reconstructions, the displayed fluid region was defined using the corresponding SR-derived PCMRA mask. Flow rate (Q, mL/s) and 95th-percentile velocity (v95, m/s) are shown over the cardiac cycle for each selected plane group, generated using TOF-derived masks across LR, HR, and super-resolved (SR) reconstructions. In the neurovascular case, four additional moyamoya-associated vessels (MMV-1--MMV-4) were analyzed within the disease-relevant region. Because these vessels were incompletely captured by TOF-derived arterial labels, super-resolved PCMRA-derived masks were used for both visualization and quantification of these vessels across LR, HR, and SR reconstructions.
}
\label{fig5}
\end{figure}

\begin{table*}[!]
\centering
\footnotesize
\caption{Artery-level peak-systolic flow ($Q_{\mathrm{peak}}$, mL/s), summarized as median (Q1, Q3) across all analysis planes along each vessel centerline. Values are shown for standard-resolution (LR), high-resolution (HR), CFD-pretrained (CFD), and pooled fine-tuned (POOL) reconstructions. Quantification was performed using TOF-derived masks for all vessels except the moyamoya-associated vessels (MMV-1--MMV-4), which were analyzed using super-resolved PCMRA-derived masks because they were incompletely captured by the TOF-derived arterial labels. The largest median flow per artery is shown in bold; boldface is descriptive and does not indicate statistical significance. Note that in the neurovascular subject, the LMCA is not reported; the affected proximal M1 segment was instead represented by the moyamoya-associated vessels.
}
\label{tab:peak_flow_stacked}
\setlength{\tabcolsep}{5pt}
\begin{tabular}{lcccc}
\toprule
Artery & LR & HR & CFD & POOL \\
\midrule
\multicolumn{5}{l}{\textbf{Intracranial}} \\
RICA & 5.25 (4.83, 5.76) & 5.73 (5.39, 6.04) & 5.45 (4.99, 5.81) & \textbf{5.93} (5.38, 6.50)  \\
LICA & 3.73 (3.34, 4.29) & 4.28 (3.92, 4.67) & 4.02 (3.77, 4.41) & \textbf{4.49} (4.24, 4.94)  \\
BA   & 3.38 (3.12, 3.87) & \textbf{4.24} (3.98, 4.38) & 3.64 (3.52, 3.94) & 4.10 (3.87, 4.46)  \\
RMCA & 2.78 (2.66, 2.99) & 2.79 (2.22, 3.02) & 3.00 (2.98, 3.09) & \textbf{3.51} (3.34, 3.67)  \\
LMCA & 1.56 (1.31, 2.03) & 0.98 (0.36, 1.41) & 1.69 (1.63, 1.90) & \textbf{2.61} (2.45, 2.95)  \\
RPCA & 1.39 (1.00, 1.56) & 1.84 (1.58, 2.23) & 1.61 (1.19, 1.78) & \textbf{1.86} (1.42, 2.21)  \\
LPCA & 1.03 (0.94, 1.12) & 1.32 (1.26, 1.43) & 1.08 (1.05, 1.16) & \textbf{1.37} (1.32, 1.41)  \\
RACA & 1.89 (1.68, 2.09) & 2.01 (1.69, 2.23) & 2.38 (1.99, 2.73) & \textbf{2.72} (2.55, 2.88)  \\
LACA & 0.09 (0.04, 0.21) & 0.45 (0.19, 0.57) & \textbf{0.73} (0.43, 0.85) & 0.31 (0.28, 0.60)  \\
\midrule

\multicolumn{5}{l}{\textbf{Extracranial}} \\
RICA & 19.15 (15.86, 19.86) & 16.39 (14.97, 17.91) & 17.74 (16.15, 18.55) & \textbf{19.50} (17.23, 20.07)  \\
LICA & \textbf{20.31} (19.00, 21.60) & 18.07 (17.15, 18.30) & 18.93 (18.14, 19.26) & 19.62 (18.60, 20.11)  \\
RVA  & 0.86 (0.73, 0.94) & 0.83 (0.69, 0.93) & 1.02 (0.93, 1.07) & \textbf{1.15} (1.08, 1.22)  \\
LVA  & 6.38 (5.99, 6.66) & 5.49 (4.76, 5.70) & 5.71 (5.53, 5.95) & \textbf{6.53} (6.36, 6.67)  \\
\midrule

\multicolumn{5}{l}{\textbf{Neurovascular}} \\
RICA  & 6.17 (5.61, 6.77) & 5.56 (5.23, 6.44) & 5.73 (5.28, 6.21) & \textbf{6.75} (6.36, 7.38)  \\
LICA  & 6.27 (5.73, 6.95) & 6.18 (5.22, 7.25) & 5.62 (5.36, 6.25) & \textbf{6.87} (6.18, 7.48)  \\
BA    & 2.40 (2.01, 2.65) & 2.15 (1.85, 2.67) & 2.04 (1.91, 2.08) & \textbf{2.79} (2.52, 3.00)  \\
RMCA  & 2.19 (1.74, 3.18) & 2.20 (1.99, 2.42) & 3.17 (3.06, 3.38) & \textbf{4.57} (4.11, 5.11)  \\
RPCA  & 1.03 (0.84, 1.20) & 1.21 (0.84, 1.40) & 1.16 (0.75, 1.41) & \textbf{1.56} (1.12, 1.67)  \\
LPCA  & 1.54 (0.35, 1.70) & 1.66 (0.80, 2.02) & 1.92 (0.56, 2.17) & \textbf{2.26} (0.79, 2.48)  \\
RACA  & 1.04 (0.92, 1.15) & 1.13 (0.89, 1.37) & 1.40 (1.30, 1.44) & \textbf{2.08} (1.99, 2.25)  \\
LACA  & 1.60 (1.37, 2.01) & 1.60 (1.36, 1.83) & 2.37 (2.14, 2.48) & \textbf{3.16} (2.85, 3.74)  \\
RPCOM & 0.47 (0.39, 0.55) & 0.54 (0.48, 0.74) & 0.53 (0.47, 0.54) & \textbf{0.64} (0.57, 0.68)  \\
LPCOM & 1.54 (1.37, 1.86) & 1.72 (1.63, 1.84) & 1.73 (1.57, 1.96) & \textbf{2.04} (1.81, 2.33)  \\

MMV-1  & 0.85 (0.56, 1.27) & 0.98 (0.84, 1.17) & 1.01 (0.79, 1.45) & \textbf{1.27} (0.96, 1.98)  \\
MMV-2  & 0.61 (0.58, 0.88) & 0.67 (0.56, 0.73) & 0.64 (0.62, 0.88) & \textbf{0.81} (0.75, 1.08)  \\
MMV-3 & 0.41 (0.41, 0.44) & 0.62 (0.43, 1.23) & 0.61 (0.59, 0.62) & \textbf{0.70} (0.67, 0.72)  \\
MMV-4 & 0.30 (0.22, 0.40) & 0.47 (0.34, 0.60) & 0.48 (0.39, 0.56) & \textbf{0.49} (0.41, 0.63)  \\

\midrule

\multicolumn{5}{l}{\textbf{Neuroinflammatory}} \\
RICA & 4.02 (3.70, 4.29) & 3.85 (3.61, 4.05) & 3.99 (3.73, 4.17) & \textbf{4.28} (4.04, 4.58)  \\
LICA & 4.40 (4.15, 4.69) & 4.34 (4.03, 4.68) & 4.31 (4.05, 4.51) & \textbf{4.69} (4.40, 4.99)  \\
BA   & 2.43 (2.27, 2.67) & 2.43 (2.23, 2.60) & 2.39 (2.24, 2.45) & \textbf{2.70} (2.55, 2.89)  \\
RMCA & 2.01 (1.66, 2.23) & 2.21 (1.95, 2.37) & 2.63 (2.54, 2.75) & \textbf{3.18} (2.95, 3.40)  \\
LMCA & 2.56 (2.52, 2.67) & 2.59 (2.38, 2.81) & 2.57 (2.52, 2.62) & \textbf{3.10} (3.01, 3.18)  \\
RPCA & 0.67 (0.55, 0.87) & 0.89 (0.79, 0.99) & 0.96 (0.90, 1.03) & \textbf{1.29} (1.20, 1.37)  \\
LPCA & 0.65 (0.47, 0.75) & 0.85 (0.70, 0.95) & 0.95 (0.91, 1.00) & \textbf{1.48} (1.25, 1.66)  \\
RACA & 0.69 (0.59, 0.89) & 0.95 (0.76, 1.13) & 1.04 (0.97, 1.14) & \textbf{1.76} (1.60, 1.94)  \\
LACA & 1.15 (1.04, 1.32) & 1.23 (1.07, 1.38) & 1.42 (1.25, 1.50) & \textbf{1.68} (1.55, 1.78)  \\
\bottomrule
\end{tabular}
\end{table*}

Figure~\ref{fig5} summarizes plane-based hemodynamic comparisons across all four test cases. Visual inspection of the velocity-field panels showed that the super-resolved reconstructions had more spatially continuous velocity patterns than LR, particularly within higher-velocity flow regions. In the neurovascular case, TOF-derived segmentation incompletely captured several smaller vessels within the moyamoya-associated region. Four such vessels were therefore selected for exploratory analysis and labeled MMV-1--MMV-4, with super-resolved PCMRA-derived masks used to define these regions consistently across LR, HR, CFD, and POOL.

The super-resolved flow-rate ($Q$) and $v_{p95}$ curves broadly followed the temporal profiles observed in the acquired LR and HR data. Both super-resolution models generally shifted plane-based $v_{p95}$ toward the HR reference, with estimates falling between LR and HR across most displayed plane groups. For flow rate, POOL produced the highest mean peak-systolic estimate across the 10 displayed plane groups (4.54 mL/s vs 4.08 mL/s for LR, 3.82 mL/s for HR, and 4.02 mL/s for CFD).

Table~\ref{tab:peak_flow_stacked} provides artery-level peak-systolic flow estimates across all analyzed vessel segments. POOL yielded the highest median peak-systolic flow ($Q_{\mathrm{peak}}$) in 33 of 36 arteries. Within the smaller intracranial vessels (MCA, PCA, ACA, PCOM, and moyamoya-associated vessels), CFD exceeded LR in 23 of 23 arteries and HR exceeded LR in 21 of 23, while CFD exceeded HR in 16 of 23. In the larger proximal vessels, differences between LR, HR, and CFD were more mixed. In the extracranial case, LR peak flow was generally higher than HR, consistent with different sampling of the systolic peak in the separately acquired scans.

\begin{figure}[!]
\centering
\includegraphics[width=1.0\textwidth]{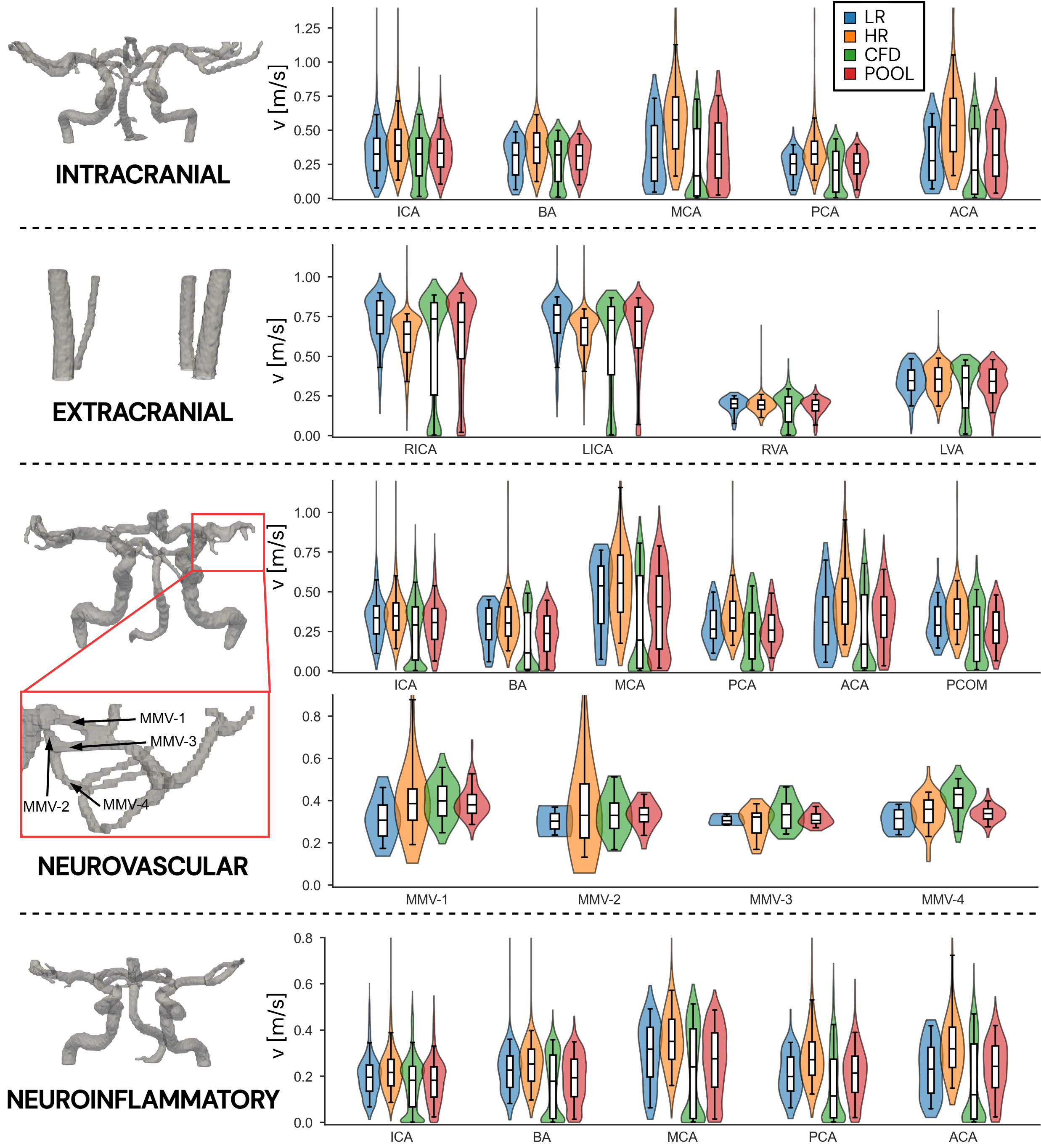}
\caption{Voxel-wise peak-systolic velocity magnitude distributions, shown as violin plots for each labeled artery and cohort. The violin plots compare standard-resolution (LR), high-resolution (HR), CFD-pretrained (CFD), and pooled fine-tuned (POOL) reconstructions. Box plots within each violin indicate the median and interquartile range. For the primary vessel analyses, voxel-wise distributions were computed within TOF-derived arterial masks, shown as vessel renderings on the left. In the neurovascular case, four additional moyamoya-associated vessels (MMV-1--MMV-4) were analyzed within the disease-relevant region using super-resolved PCMRA-derived masks, shown in the zoomed vessel rendering, because these vessels were incompletely captured by the TOF-derived arterial labels.
}
\label{fig6}
\end{figure}
\FloatBarrier

\begin{table*}[t!]
\centering
\setlength{\tabcolsep}{3pt}
\footnotesize
\caption{Artery-level velocity distribution metrics at peak systole. Median velocity ($v_{\mathrm{median}}$), reported with the first and third quartiles (Q1, Q3), and the 95th percentile ($v_{\mathrm{p95}}$) characterize the voxel-wise velocity distribution within each artery, while mean absolute divergence (div) is reported in units of $10$~s$^{-1}$. Values are shown for standard-resolution (LR), high-resolution (HR), CFD-pretrained (CFD), and pooled fine-tuned (POOL) reconstructions. Velocity values are given in cm/s. Quantification was performed using TOF-derived masks for all vessels except the moyamoya-associated vessels (MMV-1--MMV-4), which were analyzed using super-resolved PCMRA-derived masks because they were incompletely captured by the TOF-derived arterial labels. For readability, the largest value for each velocity-distribution metric and the lowest divergence value per artery are shown in bold; boldface is descriptive and does not indicate statistical significance. Note that in the neurovascular subject, the LMCA is not reported; the affected proximal M1 segment was instead represented by the moyamoya-associated vessels.
}
\label{tab:velocity_distribution_stacked}
\hspace*{-0.15\textwidth}%
\resizebox{1.3\textwidth}{!}{%
\begin{tabular}{lcccccccccccc}
\toprule
& \multicolumn{3}{c}{LR}
& \multicolumn{3}{c}{HR}
& \multicolumn{3}{c}{CFD}
& \multicolumn{3}{c}{POOL} \\
Artery
& $v_{\mathrm{median}}$ (Q1, Q3) & $v_{\mathrm{p95}}$ & div
& $v_{\mathrm{median}}$ (Q1, Q3) & $v_{\mathrm{p95}}$ & div
& $v_{\mathrm{median}}$ (Q1, Q3) & $v_{\mathrm{p95}}$ & div
& $v_{\mathrm{median}}$ (Q1, Q3) & $v_{\mathrm{p95}}$ & div \\
\midrule

\multicolumn{13}{l}{\textbf{Intracranial}} \\
RICA & 33.0 (20.9, 43.3) & 60.8 & 6.9
     & \textbf{38.6 (26.4, 50.6)} & \textbf{72.1} & 16.2
     & 32.9 (16.7, 43.6) & 61.3 & \textbf{3.3}
     & 33.0 (22.5, 43.0) & 58.5 & 4.7 \\

LICA & 32.2 (20.0, 44.7) & 62.4 & 6.8
     & \textbf{39.3 (28.5, 50.9)} & \textbf{70.7} & 17.6
     & 32.2 (16.4, 45.2) & 62.4 & \textbf{3.5}
     & 33.1 (23.4, 44.1) & 60.5 & 4.9 \\

BA   & 31.7 (17.2, 40.8) & 48.7 & 5.2
     & \textbf{37.5 (25.8, 48.1)} & \textbf{61.5} & 18.5
     & 31.9 (12.4, 41.9) & 50.0 & \textbf{3.0}
     & 31.2 (21.1, 39.3) & 47.4 & 3.7 \\

RMCA & 40.0 (16.5, 56.5) & 71.8 & 6.4
     & \textbf{58.2 (40.3, 72.5)} & \textbf{103.6} & 46.8
     & 23.5 (2.3, 52.9) & 69.0 & \textbf{4.9}
     & 32.4 (15.7, 51.5) & 67.6 & 5.6 \\

LMCA & 23.2 (9.1, 50.3) & 77.6 & 13.4
     & \textbf{52.8 (27.4, 77.9)} & \textbf{128.0} & 50.9
     & 6.9 (1.2, 47.7) & 79.4 & \textbf{6.8}
     & 32.4 (13.1, 60.0) & 86.0 & 10.0 \\

RPCA & 25.6 (17.6, 33.6) & 41.6 & 5.6
     & \textbf{33.5 (25.9, 44.1)} & \textbf{59.0} & 27.0
     & 21.0 (4.5, 35.4) & 44.3 & \textbf{3.7}
     & 26.2 (17.9, 33.6) & 40.8 & 4.8 \\

LPCA & 24.7 (17.0, 30.9) & 35.7 & 6.9
     & \textbf{31.6 (23.4, 39.0)} & \textbf{57.2} & 21.7
     & 19.7 (4.8, 33.3) & 42.7 & \textbf{3.3}
     & 25.5 (18.1, 31.8) & 36.8 & 4.8 \\

RACA & 36.1 (21.5, 55.2) & 61.3 & 10.5
     & \textbf{58.1 (41.1, 74.8)} & \textbf{111.0} & 48.1
     & 33.6 (3.5, 55.4) & 69.4 & \textbf{7.1}
     & 44.2 (29.4, 55.9) & 67.1 & 8.7 \\

LACA & 17.1 (10.5, 25.0) & 69.6 & 8.0
     & \textbf{37.0 (23.0, 67.2)} & \textbf{98.9} & 46.9
     & 10.4 (2.8, 31.5) & 63.7 & \textbf{5.7}
     & 17.7 (8.5, 25.6) & 56.8 & 6.2 \\

\addlinespace[0.8em]
\multicolumn{13}{l}{\textbf{Extracranial}} \\
RICA & \textbf{75.9 (64.3, 85.1)} & \textbf{90.1} & 7.5
     & 64.0 (52.5, 71.9) & 76.9 & 8.0
     & 73.5 (25.7, 83.8) & 88.6 & \textbf{4.0}
     & 71.4 (48.6, 83.8) & 89.9 & 5.0 \\

LICA & \textbf{76.1 (64.7, 82.4)} & \textbf{87.5} & 8.3
     & 68.2 (57.0, 74.1) & 79.9 & 8.2
     & 72.7 (38.5, 81.3) & 87.1 & \textbf{3.6}
     & 72.0 (55.4, 81.0) & 86.8 & 5.0 \\

RVA  & 20.1 (17.4, 23.1) & 25.3 & 5.0
     & 19.5 (16.6, 22.4) & 26.0 & 7.9
     & \textbf{20.3 (8.7, 24.6)} & \textbf{29.5} & \textbf{2.1}
     & 19.7 (15.7, 22.4) & 26.0 & 3.2 \\

LVA  & 34.6 (28.5, 41.5) & 48.4 & 3.8
     & 35.7 (27.9, 42.9) & \textbf{48.8} & 6.0
     & \textbf{36.5 (17.5, 44.1)} & 47.7 & \textbf{1.6}
     & 34.2 (27.1, 41.8) & 48.0 & 2.3 \\

\addlinespace[0.8em]
\multicolumn{13}{l}{\textbf{Neurovascular}} \\
RICA & 33.5 (23.6, 41.6) & 59.1 & 9.1
     & \textbf{33.7 (25.1, 41.8)} & \textbf{60.9} & 16.6
     & 29.6 (8.5, 40.7) & 57.0 & \textbf{3.1}
     & 30.4 (20.1, 39.4) & 55.2 & 4.2 \\

LICA & 33.7 (23.5, 41.0) & 56.0 & 8.2
     & \textbf{35.9 (26.7, 44.2)} & \textbf{59.3} & 19.2
     & 28.9 (5.7, 40.5) & 54.8 & \textbf{3.2}
     & 31.0 (20.0, 39.5) & 52.5 & 4.2 \\

BA   & 29.7 (19.9, 39.8) & 45.0 & 11.2
     & \textbf{30.4 (22.2, 40.6)} & \textbf{52.5} & 26.6
     & 11.6 (1.2, 37.0) & 49.2 & \textbf{3.6}
     & 23.9 (12.4, 34.9) & 44.8 & 4.9 \\

RMCA & 53.8 (30.0, 66.3) & 76.3 & 15.6
     & \textbf{55.5 (37.0, 73.2)} & \textbf{115.8} & 43.0
     & 19.5 (1.8, 60.3) & 80.8 & \textbf{5.4}
     & 40.7 (14.1, 59.9) & 78.9 & 8.4 \\

RPCA & 25.4 (20.1, 31.9) & 40.5 & 7.8
     & \textbf{31.0 (23.5, 38.7)} & \textbf{55.9} & 30.7
     & 21.0 (3.8, 31.8) & 40.7 & \textbf{3.4}
     & 24.2 (16.8, 30.0) & 39.6 & 5.7 \\

LPCA & 30.8 (21.9, 44.4) & 51.0 & 9.0
     & \textbf{36.9 (27.8, 48.2)} & \textbf{61.8} & 26.8
     & 26.3 (11.2, 46.6) & 57.4 & \textbf{4.2}
     & 29.2 (20.1, 41.2) & 52.3 & 5.6 \\

RACA & 30.8 (14.8, 42.1) & 48.6 & 17.2
     & \textbf{40.0 (26.0, 51.9)} & \textbf{92.9} & 33.8
     & 12.7 (2.0, 41.0) & 57.0 & \textbf{5.3}
     & 32.9 (20.3, 41.4) & 50.0 & 7.8 \\

LACA & 30.3 (17.3, 54.3) & 72.7 & 10.9
     & \textbf{48.0 (33.1, 64.6)} & \textbf{95.5} & 43.1
     & 21.5 (2.4, 54.5) & 71.0 & \textbf{6.7}
     & 40.0 (22.5, 54.2) & 69.1 & 9.1 \\

RPCOM & 22.2 (22.0, 24.5) & 26.0 & 6.3
      & \textbf{25.6 (21.3, 30.0)} & \textbf{40.8} & 34.6
      & 16.3 (6.7, 25.6) & 32.9 & \textbf{3.4}
      & 18.7 (13.9, 22.4) & 26.3 & 6.4 \\

LPCOM & 37.5 (27.3, 44.1) & 50.8 & 9.8
      & \textbf{40.0 (31.4, 46.9)} & \textbf{58.4} & 27.0
      & 32.2 (5.8, 44.5) & 53.6 & \textbf{5.5}
      & 32.6 (23.1, 40.7) & 49.6 & 6.2 \\

MMV-1 & 30.8 (23.3, 38.0) & 46.4 & 16.0
      & 38.6 (30.7, 45.6) & \textbf{87.9} & 29.7
      & \textbf{39.9 (32.8, 46.9)} & 55.7 & \textbf{6.9}
      & 38.1 (34.1, 42.9) & 52.8 & 10.2 \\

MMV-2 & 30.3 (26.6, 34.0) & 36.9 & 13.1
      & 33.1 (22.4, 48.0) & \textbf{124.8} & 37.9
      & 33.0 (26.8, 38.9) & 51.3 & \textbf{5.2}
      & \textbf{33.3 (30.1, 36.6)} & 43.0 & 9.7 \\

MMV-3 & 30.6 (28.8, 32.6) & 33.7 & 7.3
      & 32.3 (24.7, 34.1) & 38.6 & 17.0
      & \textbf{33.4 (27.0, 38.6)} & \textbf{46.5} & 2.8
      & 30.7 (29.3, 33.8) & 37.3 & \textbf{2.5} \\

MMV-4 & 31.6 (26.5, 35.7) & 38.3 & 20.9
      & 36.0 (29.6, 40.4) & 44.0 & 42.6
      & \textbf{43.0 (36.6, 46.0)} & \textbf{50.5} & \textbf{4.8}
      & 33.9 (31.4, 36.2) & 39.9 & 6.9 \\

\addlinespace[0.8em]
\multicolumn{13}{l}{\textbf{Neuroinflammatory}} \\
RICA & 19.0 (12.4, 24.0) & 32.9 & 10.8
     & \textbf{21.0 (15.5, 26.4)} & \textbf{38.8} & 11.9
     & 17.7 (6.5, 23.9) & 33.1 & \textbf{1.2}
     & 17.8 (10.6, 23.5) & 31.9 & 2.6 \\

LICA & 20.4 (14.7, 25.7) & 35.3 & 9.5
     & \textbf{22.3 (16.4, 28.2)} & \textbf{39.1} & 10.8
     & 18.8 (7.1, 24.9) & 35.7 & \textbf{1.3}
     & 18.9 (11.6, 24.9) & 33.9 & 2.7 \\

BA   & 22.7 (15.2, 28.9) & 36.1 & 23.6
     & \textbf{25.4 (17.9, 31.8)} & \textbf{39.9} & 18.4
     & 17.9 (1.7, 29.2) & 35.8 & \textbf{1.7}
     & 19.5 (11.2, 27.4) & 34.9 & 3.4 \\

RMCA & 34.1 (17.2, 42.9) & 51.4 & 12.3
     & \textbf{35.8 (28.2, 46.9)} & \textbf{60.6} & 25.3
     & 19.5 (1.2, 40.6) & 52.6 & \textbf{2.9}
     & 27.2 (13.3, 38.6) & 49.5 & 4.5 \\

LMCA & 30.5 (20.4, 39.2) & 47.7 & 11.4
     & \textbf{34.2 (27.0, 43.1)} & \textbf{52.5} & 15.9
     & 26.8 (3.1, 40.8) & 50.1 & \textbf{3.8}
     & 28.3 (17.5, 39.1) & 47.9 & 4.4 \\

RPCA & 22.0 (15.2, 28.7) & 33.8 & 20.4
     & \textbf{27.8 (20.9, 34.8)} & \textbf{45.6} & 26.6
     & 12.4 (2.4, 27.7) & 40.2 & \textbf{3.2}
     & 20.4 (12.7, 27.3) & 35.5 & 4.4 \\

LPCA & 19.0 (10.9, 27.3) & 36.0 & 18.1
     & \textbf{26.7 (19.7, 36.0)} & \textbf{59.2} & 26.0
     & 10.3 (1.7, 27.2) & 48.4 & \textbf{3.3}
     & 22.9 (13.8, 31.3) & 42.9 & 5.0 \\

RACA & 21.4 (10.1, 26.9) & 42.0 & 15.9
     & \textbf{31.5 (22.8, 41.6)} & \textbf{72.7} & 36.8
     & 7.7 (1.4, 28.8) & 49.4 & \textbf{4.9}
     & 25.3 (16.3, 33.7) & 42.1 & 6.1 \\

LACA & 25.9 (14.1, 33.2) & 41.9 & 20.9
     & \textbf{32.7 (24.3, 41.3)} & \textbf{71.0} & 30.3
     & 18.3 (1.9, 35.9) & 46.5 & \textbf{3.1}
     & 23.8 (13.2, 32.5) & 41.7 & 5.3 \\

\bottomrule
\end{tabular}%
}
\end{table*}

Figure~\ref{fig6} summarizes peak-systolic voxel-wise velocity distributions, with complete artery-level summary statistics provided in Table~\ref{tab:velocity_distribution_stacked}. Across the intracranial cohorts, HR generally showed higher median and 95th-percentile velocities than LR, particularly in smaller distal arteries. The opposite pattern was observed in the extracranial case, where LR velocities generally exceeded HR, consistent with the differing sampling of the systolic peak between the separately acquired scans.

CFD reconstructions showed a pronounced accumulation of near-zero velocities in several arteries, resulting in more strongly skewed distributions. POOL showed less near-zero accumulation and generally higher median velocities than CFD across intracranial territories, although this pattern was not consistent in all arteries, particularly extracranially. Voxel-wise $v_{p95}$ was slightly higher for CFD than POOL across most territories, whereas the relation between POOL and LR was mixed across arteries. This differed from the plane-based $v_{p95}$ analysis, in which POOL generally exceeded LR across the displayed plane groups.

\section{Discussion}\label{sec5}

In this study, we performed a preliminary multi-cohort \textit{in vivo} evaluation of super-resolution for intracranial 4D Flow MRI, focusing on synthetic-to-\textit{in vivo} generalization and the effect of transfer learning across neurological cohorts and acquisition settings. The synthetic-pretrained model generalized across all four test cohorts, while transfer learning provided additional improvements in reconstruction error and voxel-wise agreement. Pooled multi-cohort fine-tuning showed the most consistent overall performance compared with cohort-specific adaptation. Super-resolution also improved angiographic vessel depiction and enabled exploratory hemodynamic assessment in vessel segments that were poorly depicted at native resolution. Together, these findings support super-resolution 4D Flow MRI as a proof-of-principle approach to higher-resolution cerebrovascular flow assessment.

The synthetic-to-\textit{in vivo} performance of the CFD-pretrained model supports the use of acquisition-mimicking synthetic data for pretraining~\cite{odeback2026potential}. Previous intracranial 4D Flow MRI super-resolution studies have largely relied on synthetic training data generated through simpler downsampling or k-space cropping procedures, with \textit{in vivo} evaluation often limited to small cohorts or indirect reference settings~\cite{shit2022srflow, ferdian2023cerebrovascular, rutkowski2021enhancement}. In contrast, our pipeline produces a synthetic domain sufficiently representative of clinical image data to support evaluation without \textit{in vivo} adaptation, consistently yielding higher voxel-wise agreement as compared to the reference 4DFlowNet network~\cite{ferdian2023cerebrovascular}, and serving as a solid pretrained foundation for subsequent transfer learning.

Continuing, transfer learning incrementally improved reconstruction accuracy, revealing a practical observation: diversity in fine-tuning data may be preferred over narrowly targeted tuning. While cohort-specific models performed well within their own domain, they showed reduced cross-cohort performance under \textit{domain shift}, most notably when transferring from extracranial carotid vessels into intracranial distal arteries. In contrast, pooled multi-cohort fine-tuning consistently exceeded cohort-specific performance, even when compared to domain-specific networks. While extension into \emph{even} more distal vessels remains undetermined, these observations support further exploration of broadly adapted models trained on heterogeneous clinical cases, rather than relying exclusively on separate models for each clinical subgroup.

Beyond voxel-wise reconstruction metrics, super-resolution improved vessel depiction on derived PCMRA, with reduced background signal and clearer delineation of major arterial branches – improvements of possible relevance to downstream image analysis tasks including segmentation, centerline extraction, and standardized placement of analysis planes. In the moyamoya case, smaller vessels surrounding the affected proximal M1 region appeared more conspicuous after super-resolution and were incompletely captured by the TOF-derived arterial labels. This enabled exploratory visualization and quantification using super-resolved PCMRA masks. However, these vessels are small, tortuous, and close to or below the effective spatial resolution of the acquired data, and the high-resolution reference remains uncertain in this territory. Disease-specific conclusions therefore require validation in larger moyamoya cohorts and against independent angiographic or physiologic references.

Compared with previous cerebrovascular super-resolution studies, which have often relied on synthetic or indirect reference settings or limited \textit{in vivo} validation, the present study directly evaluated synthetic-pretrained and transfer-learned models against acquired high-resolution data across heterogeneous cohorts~\cite{rutkowski2021enhancement, shit2022srflow, ferdian2023cerebrovascular}. By combining voxel-wise reconstruction metrics with complementary angiographic and hemodynamic evaluation, the findings provide early evidence for the translational potential of intracranial super-resolution 4D Flow MRI. Although the approximately 9-minute standard-resolution protocol substantially reduced scan time relative to the high-resolution reference, it remains lengthy for routine clinical use. These findings should therefore be viewed as an intermediate step toward clinically feasible intracranial 4D Flow MRI, alongside continued work on acquisition acceleration, motion correction, reproducibility, and disease-specific hemodynamic validation~\cite{mahinrad2022intracranial, fico2023cerebral, aristova2022accelerated, cao2024observation, wen2019test, rivera2022motion, waahlin20224d, xia2025dynamic}.

This study has a few notable limitations. First, cohort sizes were modest, particularly for the disease cohorts, and only one participant per cohort was held out for final testing. The analyses were therefore descriptive and cannot establish participant-level statistical significance or disease-specific effects. The cohorts were also not demographically matched and differed in scanner platform and acquisition characteristics, which may confound direct cross-cohort comparisons, although this heterogeneity also provided a pragmatic test of domain shift. Second, voxel-wise fine-tuning and evaluation relied on HR-derived LR inputs, ensuring exact spatiotemporal correspondence; these inputs may not reproduce all artifacts present in independently acquired standard-resolution scans. Native standard-resolution acquisitions were therefore retained for angiographic and hemodynamic analyses. Third, the high-resolution scans were themselves noisy, particularly near vessel boundaries, and TOF-derived reference masks were subject to registration and thresholding uncertainty. Finally, no independent external test dataset or independent physiologic reference, such as flow-phantom or invasive flow measurements, was available. External multi-center validation and independent reference measurements will therefore be required before clinical performance can be established.

In conclusion, synthetic-pretrained deep learning-based super-resolution generalized to heterogeneous \textit{in vivo} 4D Flow MRI data, with transfer learning providing additional gains and pooled multi-cohort fine-tuning showing the most consistent overall performance. Super-resolution also improved vessel depiction and supported exploratory hemodynamic assessment in small cerebrovascular territories. These findings support further multi-center validation of broadly adapted super-resolution models for intracranial 4D Flow MRI.

\section*{Funding}
This work was supported by the European Union through the European Research Council (ERC), MultiPRESS (grant agreement No. 101075494). The views and opinions expressed are those of the authors and do not necessarily reflect those of the European Union or the European Research Council Executive Agency. This work was also funded in-part through the PIRATE project under the 2021--2027 ERDF Programme of the State of Mecklenburg-Western Pomerania by the European Regional Development Fund (ERDF) of the European Union. The funders had no role in study design, data collection, data analysis, interpretation of the results, or preparation of the manuscript.

\section*{Acknowledgments}
Computations were performed on resources provided by the National Academic Infrastructure for Supercomputing in Sweden at the National Supercomputer Centre at Linköping University (Berzelius).

During manuscript preparation, the authors used ChatGPT (GPT-5.6, OpenAI, last accessed October 16, 2026) for language editing and text refinement. All AI-assisted edits were reviewed and approved by the authors, who take full responsibility for the content of the manuscript.

\section*{Data availability}
Participant-level imaging data are not publicly available due to participant privacy and ethical restrictions. Code for model training and evaluation, together with trained model weights, will be made publicly available upon publication.

\bibliography{sn-bibliography}

\clearpage
\section*{Supplementary Material}
\setcounter{section}{0}
\setcounter{figure}{0}
\setcounter{table}{0}
\setcounter{equation}{0}
\renewcommand{\thesection}{S\arabic{section}}
\renewcommand{\thefigure}{S\arabic{figure}}
\renewcommand{\thetable}{S\arabic{table}}
\renewcommand{\theequation}{S\arabic{equation}}
\renewcommand{\theHsection}{supp.\arabic{section}}

\section{Image preprocessing, vessel labeling, analysis plane placement, and direction-consistency mask refinement}\label{S1}

This section provides implementation details for (i) four-dimensional flow magnetic resonance imaging (4D Flow MRI) preprocessing, (ii) voxel-wise artery labeling, (iii) centerline-based analysis plane placement, and (iv) direction-consistency mask refinement used prior to voxel-wise and hemodynamic analyses.

Four-dimensional flow magnetic resonance imaging (4D Flow MRI) preprocessing followed the correction and vessel-analysis workflow described in previous work~\cite{vali2019semi, aristova2019standardized, winter2024automated}, including eddy-current correction, co-registration of time-of-flight (TOF) angiography, TOF segmentation threshold refinement, centerline extraction from the TOF-derived vessel masks, and subsequent vessel analysis. Extracranial datasets were processed by an investigator with more than 10 years of experience in neurovascular 4D Flow MRI. Intracranial datasets were processed by a doctoral researcher who had been trained in the workflow by the experienced investigator and subsequently performed processing independently according to the same protocol.

TOF angiography served as a common anatomical reference across LR, HR, and super-resolved reconstructions. Although TOF-based segmentation may be affected by signal dropout and registration error, use of the same TOF-derived segmentation reduced the potential for reconstruction-dependent segmentation differences to bias comparisons.

For intracranial datasets, a U-Net-based model~\cite{bisbal2026uncertainty} generated initial voxel-wise artery labels for the Circle of Willis, which were visually reviewed and corrected when needed by the doctoral researcher using ITK-SNAP~\cite{yushkevich2006user}. Extracranial vessel territories were delineated manually. The following artery labels were used for reporting: internal carotid arteries (ICAs), middle cerebral arteries (MCAs), anterior cerebral arteries (ACAs), posterior cerebral arteries (PCAs), basilar artery (BA), and posterior communicating arteries (PCOMs) when clearly identifiable. For the extracranial cohort, labels comprised the ICAs and vertebral arteries (VAs), manually labeled by the doctoral researcher. No formal inter- or intrarater variability analysis was performed.

For plane-based hemodynamic quantification, analysis planes were placed along each labeled vessel segment using the corresponding TOF-derived centerlines. Planes were generated perpendicular to the local centerline tangent and placed at equidistant spacing of 0.5\,mm along the centerline. For the moyamoya-associated vessels (MMV-1--MMV-4), a spacing of 0.25\,mm was used to account for their smaller size. Each plane group and artery reported was visually quality-controlled to exclude planes at intersecting bifurcations, and to ensure consistent lumen coverage across reconstructions. 

\begin{figure}[t]
  \centering
    \makebox[\textwidth][c]{%
        \includegraphics[width=1.4\textwidth]{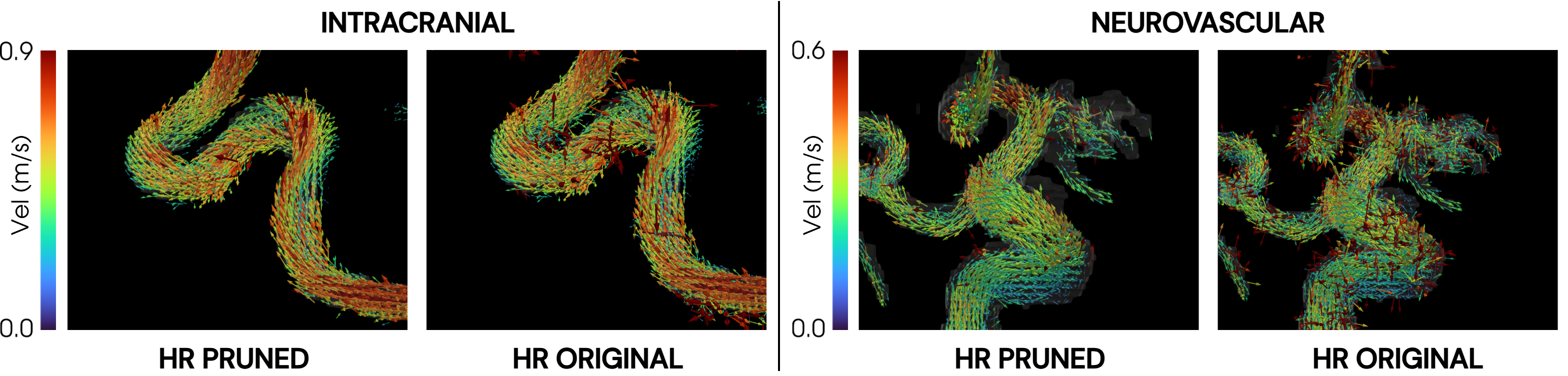}
    }
\caption{Direction-consistency mask refinement example. Velocity vectors are shown for two representative, high-resolution (HR) test sets (Intracranial and Neurovascular) after refinement (PRUNED) and before (ORIGINAL), illustrating suppression of isolated locally incoherent voxels.}
  \label{figS1}
\end{figure}

To improve robustness to isolated spurious velocity voxels - particularly prevalent near the vessel boundary - a direction-consistency refinement step was applied within the TOF-derived lumen mask prior to voxel-wise comparisons and hemodynamic analyses. Specifically, voxels were removed if their velocity direction deviated by more than an angular threshold $\theta$ from the mean direction of neighboring velocity vectors within a 5-voxel radius, with $\theta$ chosen in the range 60-90$^\circ$ depending on the dataset. A qualitative example of the refinement effect is shown in Figure~\ref{figS1}, illustrating suppression of isolated locally incoherent voxels.

\section{Network architecture, loss functions, and implementation details}

Synthetic pretraining data were generated from four patient-specific cerebrovascular computational fluid dynamics (CFD) simulations previously used in related intracranial super-resolution studies~\cite{ferdian2023cerebrovascular,odeback2026potential}. The simulations comprised time-resolved 3D velocity fields derived from patient-specific cerebrovascular geometries under physiologic boundary conditions and included anatomies with varying degrees of steno-occlusive disease. Each simulated velocity field was paired with a co-registered magnitude volume derived from clinical magnitude data and mapped to the CFD anatomy. Paired synthetic high- and low-resolution 4D Flow MRI were then generated using an acquisition-mimicking pipeline incorporating k-space truncation, noise injection matched to clinically observed levels, and dual-velocity-encoding reconstruction. This provided inherently co-registered high- and low-resolution pairs for supervised pretraining before in vivo fine-tuning. Further details are provided in~\cite{odeback2026potential}.

We implemented a 3D residual convolutional neural network (CNN) based on the 4DFlowNet-inspired~\cite{ferdian2023cerebrovascular} generator design in~\cite{odeback2026potential}, using residual-in-residual dense blocks (RRDBs)~\cite{wang2018esrgan} as the main feature extractor. The network takes as input a low-resolution velocity patch $\mathbf{v}=[u,v,w]$ and magnitude information, which are first encoded by two 3D convolution layers per branch. Velocity and magnitude features are concatenated and fused by an additional 3D convolution to form a shared low-resolution feature representation. This representation is then processed by a stack of four RRDBs in the low-resolution space. Spatial upsampling to the target grid is performed using trilinear interpolation. In the high-resolution space, features are refined by an additional stack of two RRDBs. Finally, the network branches into three output heads with two 3D convolutions each to predict the super-resolved velocity components $\hat{u}$, $\hat{v}$, and $\hat{w}$, which are concatenated to form the final 3-channel output. All convolution layers use symmetric padding.

The \textit{in silico} training objective was defined as
\begin{equation}
L_{\mathrm{in\text{-}silico}} =
\mathrm{MSE}_{\mathrm{Non\text{-}Fluid}} +
\mathrm{MSE}_{\mathrm{Boundary}} +
\mathrm{MSE}_{\mathrm{Core}} +
\mu \sum \theta^2,
\end{equation}
where the mean-squared error (MSE) was computed over three disjoint regions
$\Omega \in \{\mathrm{Non\text{-}Fluid}, \mathrm{Boundary}, \mathrm{Core}\}$.
Here, $\mathrm{Boundary}$ and $\mathrm{Core}$ denote near-wall and core-flow voxels, respectively, obtained by morphological erosion of the binary vessel mask. For a given region $\Omega$, the MSE was computed as
\begin{equation}
\mathrm{MSE}_{\Omega} = \frac{1}{N_\Omega}\sum_{i \in \Omega}
\left\lVert \mathbf{x}^{i}_{\mathrm{SR}} - \mathbf{x}^{i}_{\mathrm{HR}} \right\rVert_2^2,
\label{eq:mse_region}
\end{equation}
where $\mathbf{x}^{i}_{\mathrm{SR}}$ and $\mathbf{x}^{i}_{\mathrm{HR}}$ denote the predicted super-resolved and reference high-resolution velocity vectors at voxel $i$, and $N_\Omega$ is the number of voxels in region $\Omega$. The final term, $\mu\sum \theta^2$, denotes an $\ell_2$ weight decay with coefficient empirically set to \(\mu = 5 \cdot 10^{-7}\).

For \textit{in vivo} fine-tuning, spatially matched standard- (LR) and high-resolution (HR) patch sets were generated from the HR acquisitions, with LR inputs obtained by k-space cropping. Paired LR and HR patches were then extracted using the same patch-based sampling strategy as in the synthetic setting. Because \textit{in vivo} HR data are affected by noise and may contain isolated spurious voxels, particularly near vessel boundaries, the fine-tuning objective was simplified to separate only non-fluid and fluid regions:
\begin{equation}
L_{\mathrm{in\text{-}vivo}} =
\mathrm{MSE}_{\mathrm{Non\text{-}Fluid}} +
\mathrm{MSE}_{\mathrm{Fluid}} +
\mu \sum \theta^2 .
\end{equation}
The non-fluid region was defined as voxels outside the full co-registered TOF-derived vessel mask. Additionally, to further reduce the influence of noisy high-resolution target voxels within the vessel, the fluid region was defined using a refined vessel mask based on the aforementioned voxel removal strategy (Section~\ref{S1}) for each case, designed to exclude spurious voxels while delineating the lumen. This formulation avoids training on unreliable HR target voxels, while preserving the denoising behavior learned during synthetic pretraining.

All models were implemented in TensorFlow (v2.10) using Python and trained on an NVIDIA A100 Tensor Core GPU (40 GB). The framework comprised one synthetic-pretrained baseline model and five transfer-learned variants.
For synthetic pretraining, the network was initialized using a Glorot uniform initializer and trained for 100 epochs using the Adam optimizer with batch size 20 and an initial learning rate of $1\times10^{-4}$, decayed by a factor of 2 after 50 epochs. Transfer-learned models were initialized from the synthetic-pretrained weights and fine-tuned using Adam with a learning rate of $1\times10^{-5}$ while freezing the initial low-resolution feature-extraction blocks (four RRDBs). Freezing the early layers was intended to reduce the number of trainable parameters and limit over-adaptation to the small target dataset and was selected based on preliminary model-development experiments. Because a formal ablation of freezing strategies was outside the scope of this study, this approach should be considered a fixed fine-tuning design choice rather than a systematically optimized strategy. Final model checkpoints were selected based on validation set performance. Total training time was approximately 19 hours for synthetic pretraining and approximately 1 hour for fine-tuning each model.

\section{Data partitions and training details}

The baseline model was pretrained on the full set of $n=4$ synthetic computational fluid dynamics (CFD)-derived velocity fields used in previous works~\cite{ferdian2023cerebrovascular, odeback2026potential}. In total, the synthetic training set comprised 123,500 aligned HR/LR patch pairs, generated by sampling 10 patches per CFD time frame and applying rigid rotation augmentation to each patch.

For \textit{in vivo} transfer learning, cohort-specific and pooled fine-tuning used the participant-level partitions summarized in Table~\ref{tab:training_splits}. One participant per cohort was held out for final testing. For the healthy intracranial and extracranial cohorts, four participants were used for training and one separate participant for validation. For the neurovascular and neuroinflammatory cohorts, two participants were used for training, with model-optimization patches resampled from one training participant because no additional participant was available. Final test partitions were participant-disjoint in all cohorts and were not used during training or model selection.

\begin{table}[b!]
\centering
\normalsize
\caption{Training, validation, and test splits for \textit{in vivo} transfer learning.}
\label{tab:training_splits}

\begin{tabular}{lcccc}
\toprule
Cohort & Total, n & Training & Validation & Test \\
\midrule
Intracranial healthy
& 6 & 4 participants & 1 participant & 1 participant \\

Extracranial healthy
& 6 & 4 participants & 1 participant & 1 participant \\

Neurovascular
& 3 & 2 participants & Patch-level$^{*}$ & 1 participant \\

Neuroinflammatory
& 3 & 2 participants & Patch-level$^{*}$ & 1 participant \\
\bottomrule
\end{tabular}

\begin{flushleft}
\small
$^{*}$Validation patches were resampled from one training participant because no separate validation participant was available.
\end{flushleft}

\end{table}

The held-out test subjects were: intracranial healthy (male, 28\,years), extracranial healthy (female, 23\,years), neuroinflammatory (male, 39\,years, multiple sclerosis), and neurovascular (female, 31\,years, with left-sided moyamoya vasculopathy involving the proximal M1 segment).

To emphasize the impact of data diversity rather than the number of samples, we adjusted the sampling rate across cohorts to account for differences in the number of subjects and cardiac frames. Specifically, cohort-specific fine-tuning used 40 patches per time frame for the intracranial and extracranial cohort models and 80 patches per time frame for the two patient cohort models, which after augmentation corresponded to 10,000 (intracranial), 15,200 (extracranial), and 8,000 patches each (neurovascular and neuroinflammatory). The multi-cohort model sampled 10 patches per time frame from intracranial and extracranial cohorts and 20 patches per time frame from each patient cohort, yielding 10,300 augmented patches in total. For validation, 50 random patches per time frame were sampled from the validation subjects across cohorts.

No additional independent external validation cohort was used. Cohort-specific models were additionally tested across cohorts to characterize cross-domain robustness.

\section{Outcome measures, reference standards, and evaluation metrics}\label{appS4:metrics}

Voxel-wise reconstruction accuracy was quantified within the TOF-derived, direction-consistency-refined fluid mask described in Section~\ref{S1}, using vNRMSE, defined as
\begin{equation}
\mathrm{vNRMSE} = \frac{1}{\max_i \left\| \mathbf{v}_{\mathrm{HR}} \right\|_2} \sqrt{\frac{1}{N}\sum_{i=1}^{N} \left\| \mathbf{v}_{\mathrm{SR}}^i - \mathbf{v}_{\mathrm{HR}}^i \right\|_2^2},
\label{eq:error_metrics}
\end{equation}
where $\mathbf{v}_{\mathrm{SR}}^i$ and $\mathbf{v}_{\mathrm{HR}}^i$ denote the super-resolved and reference velocity vectors at voxel $i$, respectively, and $N$ is the number of evaluated fluid voxels.

For Bland--Altman analysis at peak systole, differences were defined as $V_{\mathrm{SR}}-V_{\mathrm{HR}}$ and plotted against the corresponding pairwise mean. Mean bias and 95\% limits of agreement were calculated as the mean difference and mean difference $\pm1.96$ standard deviations, respectively.

To quantify angiographic vessel depiction, phase-contrast magnetic resonance angiography (PCMRA) volumes derived from LR, HR, and super-resolved velocity fields were compared against their corresponding co-registered TOF-derived binary vessel masks. TOF was used because it provided an independently acquired angiographic representation that was not generated from the tested 4D Flow MRI reconstruction. For this analysis, the native \textit{in vivo} standard-resolution acquisitions served as input to the network models. 

Evaluation was restricted to the minimal bounding box enclosing the nonzero TOF mask, and PCMRA intensities were max-normalized as $I_{\mathrm{PCMRA}} = I/\max(I)$. Because vessel voxels represented a minority class, threshold-independent vessel discrimination was summarized using the area under the precision--recall curve (PR-AUC). Within the evaluation region, normalized PCMRA intensity was treated as a continuous prediction score and the TOF-derived vessel mask as the binary reference. Precision and recall were defined as
\begin{equation}
\mathrm{Precision} = \frac{\mathrm{TP}}{\mathrm{TP}+\mathrm{FP}},
\qquad
\mathrm{Recall} = \frac{\mathrm{TP}}{\mathrm{TP}+\mathrm{FN}},
\end{equation}
where TP, FP, and FN denote true positives, false positives, and false negatives, respectively. Precision--recall curves were generated across the full range of unique PCMRA intensity thresholds, and PR-AUC was calculated using trapezoidal integration after ordering points by increasing recall:
\begin{equation}
\mathrm{PR\text{-}AUC}
=
\sum_{j=1}^{K-1}
\frac{\mathrm{Precision}_{j+1}+\mathrm{Precision}_{j}}{2}
\left(
\mathrm{Recall}_{j+1}-\mathrm{Recall}_{j}
\right).
\label{eq:prauc}
\end{equation}

Spatial overlap was assessed separately using the maximum Dice coefficient. Binary PCMRA masks were generated using
\[
I_{\mathrm{PCMRA}} > \mu + \tau\sigma,
\]
where $\mu$ and $\sigma$ denote the mean and standard deviation of the normalized PCMRA volume. The threshold parameter $\tau$ was swept from $-0.6$ to $6.0$ in increments of $0.1$, and Dice overlap was calculated at each threshold as
\begin{equation}
\mathrm{Dice}(\tau)
=
\frac{2\,\mathrm{TP}(\tau)}
{2\,\mathrm{TP}(\tau)+\mathrm{FP}(\tau)+\mathrm{FN}(\tau)}.
\label{eq:dice}
\end{equation}
The maximum value across the threshold sweep was reported as
\begin{equation}
\mathrm{Dice}_{\max}
=
\max_{\tau}\mathrm{Dice}(\tau).
\label{eq:dice_max}
\end{equation}

Finally, the divergence of the 3D velocity field was computed at peak systole as
\begin{equation}
\nabla\cdot\mathbf{v} = \frac{\partial u}{\partial x} + \frac{\partial v}{\partial y} + \frac{\partial w}{\partial z},
\label{eq:div_def}
\end{equation}
with finite differences implemented via numerical gradients along each spatial axis. Divergence was summarized within the same TOF-derived, direction-consistency-refined fluid mask used for quantitative evaluation as

\begin{equation}
\overline{\left|\nabla\cdot\mathbf{v}\right|} = \frac{1}{N}\sum_{i=1}^{N} \left|\left(\nabla\cdot\mathbf{v}\right)^i\right|,
\label{eq:div_mean_abs}
\end{equation}
where $N$ denotes the number of evaluated fluid voxels. Here, LR velocity fields were first interpolated to the HR grid before computing divergence to ensure matched spatial resolution.

Analyses were performed using Python (v3.8).

\section{4DFlowNet benchmark implementation}

As an external benchmark, we evaluated the cerebrovascular 4DFlowNet~\cite{ferdian2023cerebrovascular} using the publicly available pretrained model weights from the authors' repository (\url{https://github.com/EdwardFerdian/4DFlowNet}). The model was not retrained on the present synthetic or \textit{in vivo} datasets. For voxel-wise comparison, 4DFlowNet was applied to the same HR-derived LR inputs used for evaluation of the present models, and outputs were evaluated against the same HR references using the same TOF-derived vessel masks.

While complete network architecture and training details are provided in~\cite{ferdian2023cerebrovascular}, the main differences relative to the present baseline model are summarized here. First, while the same set of $n=4$ cerebrovascular flow simulations was utilized, the publicly available cerebrovascular 4DFlowNet weights were trained using the data split and preprocessing strategy described in~\cite{ferdian2023cerebrovascular}. Here, two CFD-derived cases were used for training, one for validation, and one for testing. In contrast, the present baseline model was pretrained using all four available CFD-derived cerebrovascular cases, with subsequent evaluation performed only on the held-out \textit{in vivo} test subjects described in the main manuscript. Second, the synthetic image-generation pipelines differed. In 4DFlowNet, low-resolution inputs were generated using the original cerebrovascular 4DFlowNet downsampling procedure, designed to mimic spatial-resolution degradation and noise in 4D Flow MRI. In the present work, synthetic pretraining used a related but updated simulation-to-MRI pipeline described in~\cite{odeback2026potential}, which extends this framework by incorporating a more acquisition-mimicking reconstruction procedure, including dual-velocity-encoding reconstruction. 

These differences were not modified or harmonized, because the purpose of the benchmark was to compare the present synthetic-pretrained and transfer-learned models against an established, previously published cerebrovascular super-resolution framework. Consequently, differences in performance between 4DFlowNet and the present models may reflect not only network architecture, but also differences in training-data usage and synthetic data generation.

This comparison represents benchmarking against an externally developed model, not validation on an independent external dataset.

\begin{figure}[t!]
\centering
\includegraphics[width=\linewidth]{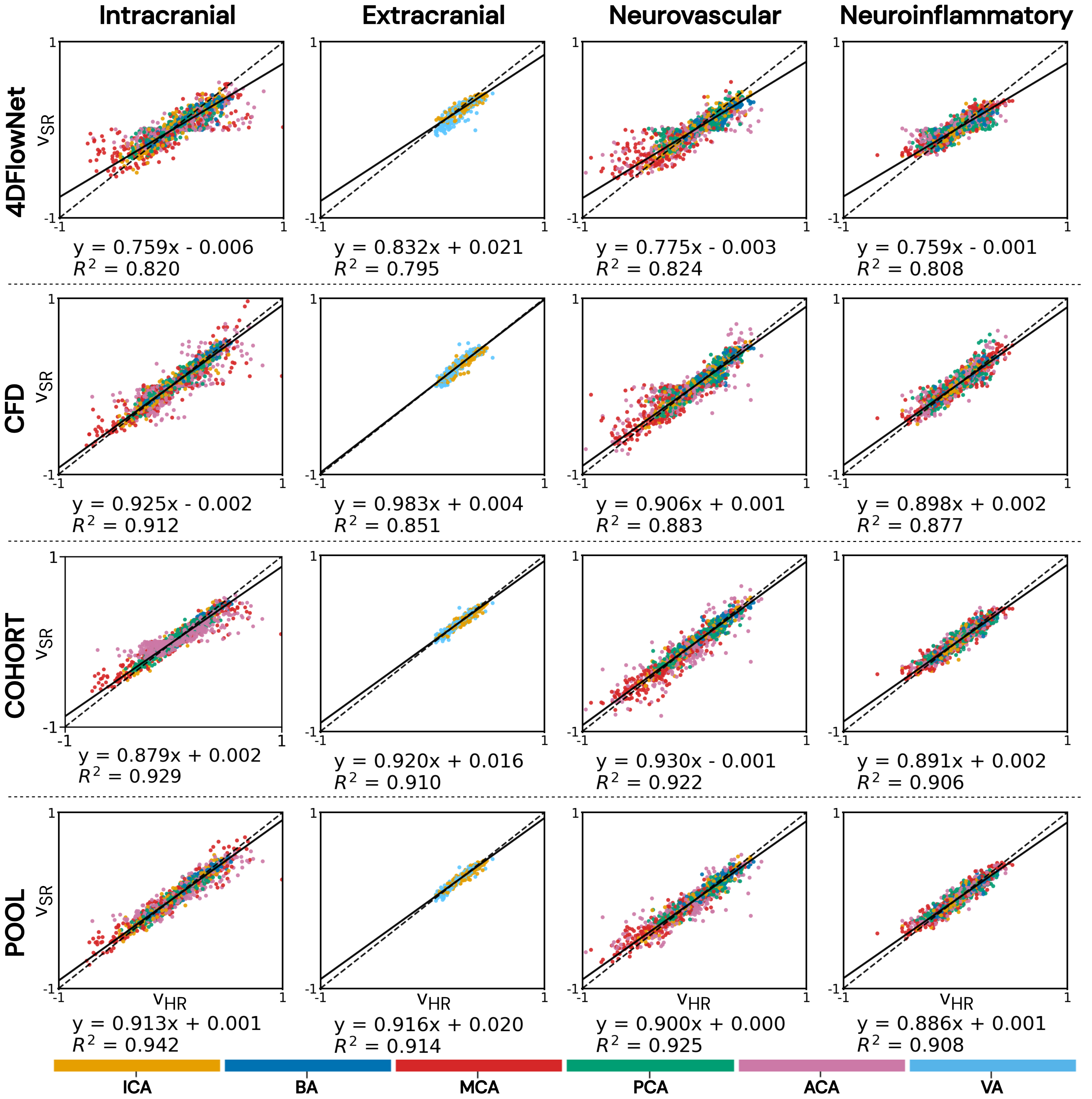}
\caption{Voxel-wise linear regression between high-resolution (HR) reference velocities (\(V_{\mathrm{HR}}\)) and super-resolved velocities (\(V_{\mathrm{SR}}\)) at peak systole for 4DFlowNet~\cite{ferdian2023cerebrovascular}, the CFD-pretrained baseline (CFD), the cohort-specific fine-tuned model (COHORT), and the pooled fine-tuned model (POOL) across the four test cases. For the intracranial, neurovascular, and neuroinflammatory cases, signed velocity components (\(V_x,V_y,V_z\)) were pooled; for the extracranial case, only \(V_z\) was analyzed, representing the predominant velocity component. Solid lines indicate fitted linear regressions and dashed lines the identity line. Regression metrics were calculated using all analyzed voxels within the vessel mask. For visualization, an equal number of voxels per artery was sampled. Points are colored according to artery label.
}
\label{figS2}
\end{figure}

\section{Voxel-wise Linear Regression}

To complement the Bland--Altman agreement analysis presented in the main manuscript, voxel-wise linear regression was performed to characterize linear association and scaling between predicted super-resolved velocities and the high-resolution reference at peak systole. Figure~\ref{figS2} summarizes the corresponding regression slopes and coefficients of determination ($R^2$) across all test cases and model variants.

Across the four test cases, 4DFlowNet yielded regression slopes of $0.759$--$0.832$ and $R^2$ values of $0.795$--$0.824$. Corresponding values were $0.898$--$0.983$ and $0.851$--$0.912$ for the CFD-pretrained model, $0.879$--$0.930$ and $0.906$--$0.929$ for the cohort-specific models, and $0.886$--$0.916$ and $0.908$--$0.942$ for the pooled model.
\clearpage
\end{document}